\documentclass{article}

\usepackage{microtype}
\usepackage{graphicx}
\usepackage{booktabs} 

\usepackage{hyperref}
\usepackage{subfig}

\usepackage[preprint]{icml2026}

\usepackage{amsmath}
\usepackage{amssymb}
\usepackage{mathtools}
\usepackage{amsthm}
\usepackage{booktabs}
\usepackage{multirow}

\usepackage[capitalize,noabbrev]{cleveref}

\theoremstyle{plain}
\newtheorem{theorem}{Theorem}[section]

\theoremstyle{definition}

\theoremstyle{remark}

\usepackage{tikz}
\usetikzlibrary{arrows.meta, positioning, calc}
\usetikzlibrary{arrows.meta, positioning, calc, decorations.pathreplacing}
\usepackage[textsize=tiny]{todonotes}
\usepackage{enumitem}  
\newtheorem{hypothesis}[theorem]{Hypothesis}

\icmltitlerunning{Circuit Fingerprints: How Answer Tokens Encode Their Geometrical Path}

\begin{document}

\twocolumn[
\icmltitle{Circuit Fingerprints: How Answer Tokens Encode Their Geometrical Path}





\begin{icmlauthorlist}
\icmlauthor{Andres Saurez}{aff1}
\icmlauthor{Neha Sengar}{aff1}
\icmlauthor{Dongsoo Har}{aff1}
\end{icmlauthorlist}
\icmlaffiliation{aff1}{Korea Advanced Institute of Science and Technology, Daejeon 34051, South Korea}

\icmlcorrespondingauthor{Dongsoo Har}{dshar@kaist.ac.kr}

\icmlkeywords{Machine Learning, ICML}

\vskip 0.3in
]



\printAffiliationsAndNotice{}  

\begin{abstract}
Circuit discovery and activation steering in transformers have developed as 
separate research threads, yet both operate on the same representational space. 
Are they two views of the same underlying structure? We show they follow a 
single geometric principle: answer tokens, processed in isolation, encode the 
directions that would produce them. This \emph{Circuit Fingerprint} hypothesis enables circuit discovery without gradients or causal intervention---recovering comparable structure to gradient-based methods through geometric alignment alone. We validate this on standard benchmarks (IOI, SVA, MCQA) across four model families, achieving 
circuit discovery performance comparable to gradient-based methods. The same 
directions that identify circuit components also enable controlled 
steering---achieving 69.8\% emotion classification accuracy versus 53.1\% for 
instruction prompting while preserving factual accuracy. Beyond method 
development, this read-write duality reveals that transformer circuits are 
fundamentally geometric structures: interpretability and controllability 
are two facets of the same object.
\end{abstract}

\section{Introduction}

Mechanistic interpretability seeks to understand neural networks by identifying 
the components responsible for specific computations. Two approaches have emerged: \emph{circuit discovery}, which locates important components through causal intervention or gradient approximation \citep{meng2022locating, wang2023interpretability, nanda2023attribution, kramar2024atp}, and \emph{activation steering}, which controls model behavior by intervening along learned directions \citep{turner2023activation, zou2023representation}. Despite operating on the same components in the same representational space, these approaches have been studied separately.

\begin{figure}[t]
\centering
     \includegraphics[width=\linewidth]{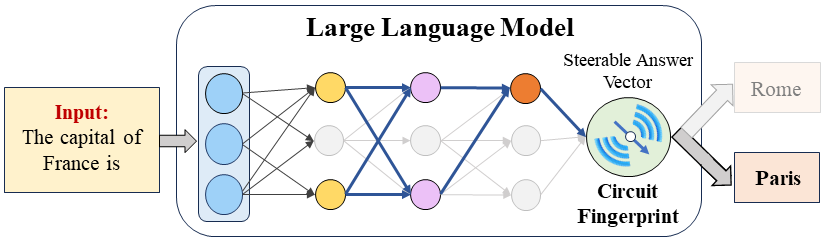}
     \caption{\textit{Circuit Fingerprints} unifies circuit discovery and activation steering as dual operations—reading and writing—on the same geometric structure encoded in answer token representations.}
     \label{fig:placeholder}
\end{figure}

We show they are two views of the same underlying structure. Answer tokens---the 
outputs a model produces---encode not just semantic content but the computational 
pathways that generated them (Fig. \ref{fig:placeholder}). The representation of ``Paris'' carries a geometric 
signature of the circuit that would produce it, regardless of context. We find 
that this signature can be \emph{read} to identify circuit membership or 
\emph{written} to steer model behavior---the same directions accomplish both, 
confirming they reflect genuine computational structure.

This finding connects circuit analysis to the linear representation hypothesis 
\citep{park2024linear, elhage2022toy}: if features are directions in activation 
space, then circuits manipulating those features should be identifiable through 
directional alignment. Our results provide direct evidence for this connection, 
suggesting that transformer circuits are fundamentally geometric structures 
encoded in the model's activation space.

We validate this hypothesis on three tasks: Indirect Object Identification (IOI)~\citep{wang2023interpretability}, 
Subject-Verb Agreement (SVA)~\citep{marks2024sparse}, and Multiple-Choice Question Answering (MCQA)~\citep{mueller2025mib}. Our geometric method is comparable ot gradient-based circuit structure without backpropagation, and the same directions enable steering that outperforms instruction prompting in emotion steering (69.8\% vs 53.1\% emotion accuracy) while preserving factual grounding. This unification simplifies the interpretability toolkit: rather than separate methods for discovery and control, a single geometric analysis reveals both which components matter and how to manipulate them.

\paragraph{Contributions.}
\begin{enumerate}
    \item We show that circuit membership can be read directly from geometric 
    alignment with answer token directions, recovering comparable circuit structure to gradient-based methods without requiring backpropagation.
    
    \item We demonstrate that the same directions enable controlled steering—read and write are dual operations on the same geometric structure—validating that we manipulate genuine computational pathways rather than superficial correlations.
    
    \item We show that feature circuits can be discovered by extracting directions from instruction-modified prompts,  without task-specific datasets.
\end{enumerate}



\section{Related Works}
\label{sec:related_works}

\paragraph{Circuit Discovery.} Identifying sparse subnetworks responsible for specific behaviors is central to mechanistic interpretability~\citep{elhage2021mathematical, olsson2022context}. The activation patching estimates causal importance by replacing activations with counterfactual values~\citep{vig2020investigating, meng2022locating, wang2023interpretability}. While providing strong causal guarantees, this requires $O(LH)$ forward passes for systematic discovery. Attribution patching~\citep{nanda2023attribution} and its integrated gradients variant EAP-IG~\citep{syed2024attribution} reduce cost via gradient approximation ~\citep{mueller2025mib}. However, gradient-based methods can suffer from saturation and sensitivity to LayerNorm non-linearities~\citep{kramar2024atp, hanna2024faith}. Optimization-based approaches frame discovery as mask learning~\citep{conmy2023towards, bhaskar2024finding}, scaling to larger models but requiring iterative optimization.

\paragraph{Activation Steering.} Representation engineering methods control model behavior by adding learned vectors to activations~\citep{turner2023activation, rimsky2024steering, zou2023representation}. These methods require contrastive data collection, direction learning, and intervention tuning. Recent work studies steering at specific circuit locations~\citep{todd2023function}, but the connection between where to steer and which directions to use remains underexplored.

\paragraph{Linear Representations.} The linear representation hypothesis posits that semantic features correspond to directions in activation space~\citep{park2024linear, elhage2022toy}, with empirical support across spatial, temporal, and conceptual features~\citep{gurnee2023language, nanda2023emergent}. Building on this, we show that if task features are geometrically encoded, their generating circuits can be found via alignment and controlled via the same directions.

\section{The Circuit Fingerprint Hypothesis}
\label{sec:circuit_fingerprint}
Circuits in transformers are not constructed dynamically for each input. 
They are stable structures encoded in the model's weights, activated 
whenever relevant computation is required \citep{olsson2022context, 
wang2023interpretability}. The same components that identify indirect 
objects operate whether the names are ``Mary and John'' or ``Alice and 
Bob.'' The same components that recall capitals operate whether the 
query concerns France or Germany.

This stability has a surprising implication. Consider two prompts:

\textit{``The answer is Paris. The capital of Italy is \_\_\_''}\\
\textit{``The capital of France is \_\_\_''}

In the first prompt, ``Paris'' appears as an input token unrelated to 
the query. Yet as it passes through the model, it traverses the same 
computational pathways responsible for \emph{producing} ``Paris''---and 
may even suppress the model from outputting ``Rome'' as the answer. 
The circuit for capital-city recall exists in the weights; ``Paris,'' 
whether as input or output, engages this structure.

\begin{hypothesis}[Circuit Fingerprint]
\label{hyp:fingerprint}
Answer tokens, processed in isolation, trace the computational 
pathways that produce them. A component belongs to the circuit 
distinguishing $a^+$ from $a^-$ if and only if processing these 
tokens activates the component differentially.
\end{hypothesis}

Following work on linear probes \citep{alain2017understanding, belinkov2022probing}, 
steering vectors \citep{turner2023activation, zou2023representation}, 
and sparse autoencoders \citep{bricken2023monosemanticity, cunningham2023sparse}, 
we assume that features are linearly encoded in activation space 
\citep{park2024linear, elhage2022toy}.

\subsection{Circuits as Computational Graphs}
\label{sec:circuits}

Transformer computation forms a directed acyclic graph where nodes 
are model components (attention heads and MLPs) and edges represent 
information flow through the residual stream. A \emph{circuit} is the 
subgraph causally responsible for a given task \citep{elhage2021mathematical, 
conmy2023towards}. Traditionally, circuit membership is established through 
intervention: a component belongs if ablating it degrades task performance 
\citep{meng2022locating, wang2023interpretability}.

The Circuit Fingerprint Hypothesis suggests an alternative characterization: 
circuits are not merely causal structures but \emph{geometric} ones, encoded 
in the linear structure of activation space. If correct, circuit membership 
can be determined from inner products with answer token directions rather than 
from interventions or gradient approximations. 


We validate this hypothesis along two axes: \textit{circuit recovery} 
(do geometric scores approximate causal importance?) and \textit{causal 
control} (can the extracted directions steer model outputs?).

\section{Reading: Geometric Attribution }
\label{sec:method}

The Circuit Fingerprint Hypothesis implies a simple approach to circuit 
discovery: if answer tokens trace the same pathways that produce them, 
then the directions separating contrastive answers should identify 
circuit components. We extract these directions, then measure which 
components align with them.

\subsection{Extracting Target and Prompt Directions}

Our method uses two sources of information: \emph{answer tokens} and 
\emph{contrastive prompts}. Answer tokens ($a^+$, $a^-$) are processed 
in isolation to extract target directions—the geometric signature of 
what distinguishes correct from incorrect outputs. Contrastive prompts 
(clean and corrupted, e.g., ``The capital of France is'' vs ``The capital 
of Italy is'') are processed to extract activation differences—how 
the model's internal state changes between conditions. Circuit components 
are identified by measuring alignment between these two: components whose 
prompt-induced activation differences align with answer-derived target 
directions are those contributing to the task.
 
We use a contrastive approach to isolate the features associated with any pair of answers and their prompts. Given answer tokens $a^+$ and $a^-$—e.g., 
``Paris'' and ``Rome''—we pass each through the model in isolation 
and compute their difference at the last layer:
\begin{equation}
\Delta \mathbf{r}^{(L)} = \mathbf{r}^{(L)}_{a^+} - \mathbf{r}^{(L)}_{a^-}
\label{eq:target_direction}
\end{equation}
This difference $\Delta \mathbf{r}^{(L)}$ defines the direction in 
activation space that distinguishes the two answers at the last layer $L$. 
Any component contributing to the distinction must write along this 
direction; any component orthogonal to it is irrelevant to the task.

Two forward passes suffice to extract target directions at all layers. 
For finer-grained analysis, we also extract channel-specific directions 
in the native spaces of each attention component:
\begin{align}
\label{eq:compo_diff}
\Delta \mathbf{v}^{(\ell,h)} &= \mathbf{v}^{(\ell,h)}_{a^+} - \mathbf{v}^{(\ell,h)}_{a^-}
& & \text{(value vectors)} \\[0.3em]
\Delta \mathbf{q}^{(\ell,h)}, \Delta \mathbf{k}^{(\ell,h)} &= \ldots
& & \text{(query/key vectors)}
\end{align}

The 3 vector differences are obtained from the prompts as this will be used later to measure the relative importance of edges towards the inputs.

\subsection{Measuring Component Direct Contributions}
\label{sec:method:contributions}

\begin{figure}
    \centering
    \includegraphics[width=0.90\linewidth]{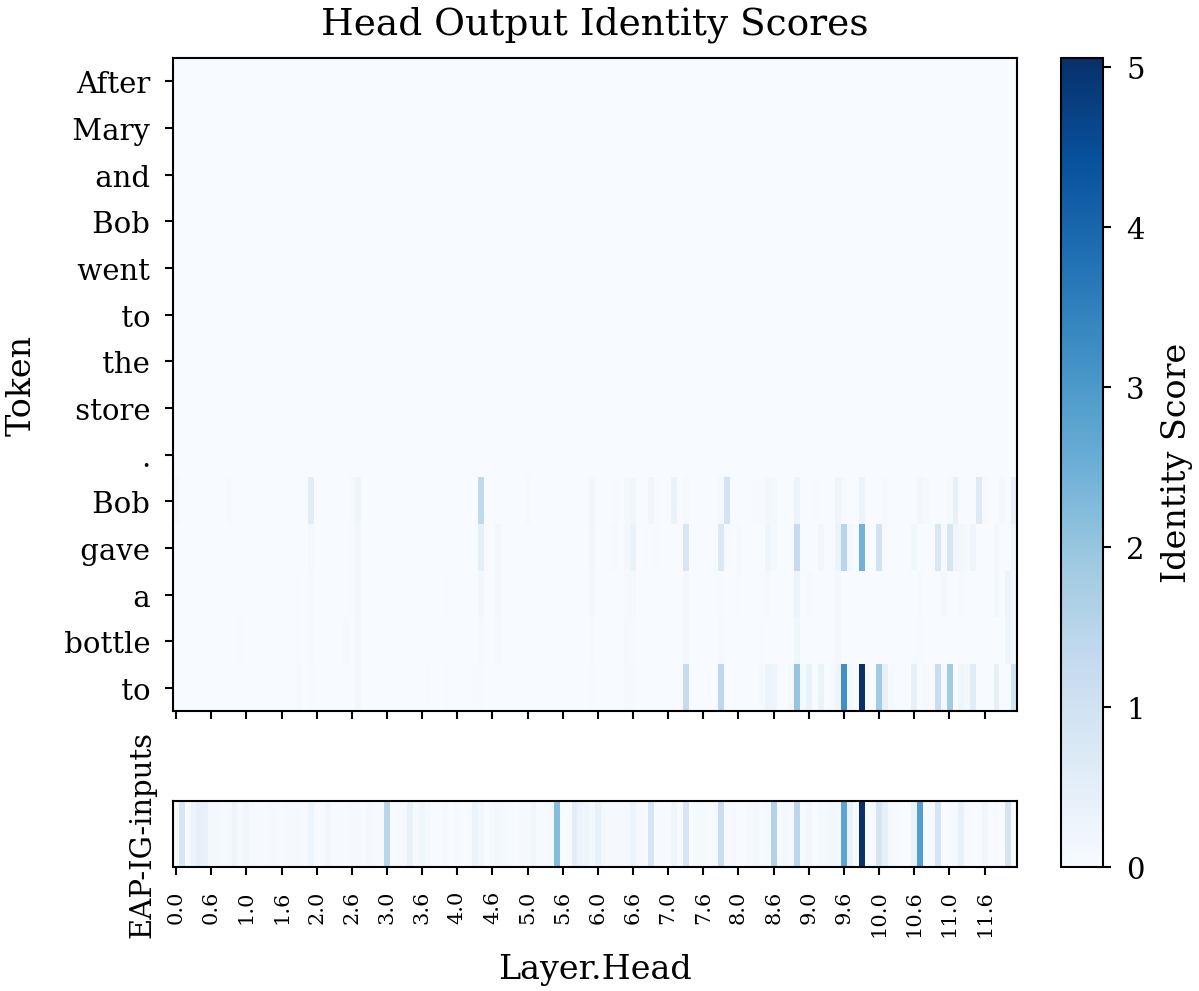}
    \caption{Comparison of attention head importance on the IOI task. \textbf{Top:} Per-token identity scores computed with respect to answer token's attention head outputs (our method). \textbf{Bottom:} Head importance from EAP-IG-inputs (gradient-based). Both methods identify the same critical heads in layers 9--11, with strongest signal at the final token position.}
    \label{fig:fig_heads_vs_circuit}
\end{figure}

The target direction $\Delta \mathbf{r}^{(L)}$ captures the features necessary 
to distinguish the two answers. We use this direction to measure how much 
each component contributes to the answer distinction---without requiring 
interventions or gradient computation.

A na\"ive approach projects each component's output into the residual stream 
and measures alignment with $\Delta \mathbf{r}^{(L)}$. However, this introduces 
noise: the projection matrices ($W_O$ for attention, $W_{out}$ for MLPs) mix 
the component's internal computation with the shared residual stream geometry.

Instead, we measure alignment in each component's \emph{native space}. For a 
component $c$ with projection matrix $W_c$ (where $W_c = W_O$ for attention 
heads and $W_c = W_{out}$ for MLPs), we first transform the target direction 
into the component's native space:
\begin{equation}
\hat{\mathbf{t}}_c = \frac{W_c^\top \Delta \mathbf{r}^{(L)}}{\| \Delta \mathbf{r}^{(L)}\|}
\label{eq:target_transform}
\end{equation}
We measure each component's task-relevant contribution as its alignment 
with this transformed target:

\begin{equation}
S_c = \left\langle \Delta \mathbf{o}_c, \; \hat{\mathbf{t}}_c \right\rangle
\label{eq:contribution}
\end{equation}
where $\Delta \mathbf{o}_c$ is the differential output of component $c$ when 
processing the contrastive prompts. Critically, this decomposition preserves 
additivity: $\sum_c S_c$ equals the total projection onto the target direction in the residual stream space, 
ensuring component contributions sum to the full effect.

As an initial observation of the contribution of this projections, consider Fig.~\ref{fig:fig_heads_vs_circuit}. For the IOI prompt ``After Mary and Bob went to the store. Bob gave a bottle to'', we display the identity scores computed by comparing attention head outputs against the answer token direct output difference. The bottom row shows head importance from EAP-IG-inputs for comparison. Both methods identify the same critical heads in layers 9--11. Interestingly, this simple analysis also reveals that the token ``gave'' exhibits a similar activation structure to the final token ``to'', suggesting that the model had as an option to output "Mary" after it. This illustrates how answer token direction encodes their own generation.

\subsection{Edge-Level Circuit Discovery}
\label{sec:method:edges}

Node-level scores (Equation~\ref{eq:contribution}) capture a component's 
\emph{direct} importance---its contribution to the final output. However, 
a component's \emph{total} importance includes its influence on downstream 
components. A head may have low direct importance but be critical because 
it shapes the queries or keys of a later head that does the heavy lifting. 
Similarly, an MLP may matter primarily through how downstream components 
read its output.

To capture this, we analyze edges: the pathways through which one component 
influences another. The residual stream at any layer is the sum of all 
upstream component outputs. We can therefore decompose a downstream 
component's task-relevant input into contributions from each upstream source.

For a target attention head $j$, information arrives through three channels 
(Q, K, V). Consider the key channel: the answer token difference 
$\Delta \mathbf{k}^{(j)}$ in key space can be projected back to residual 
stream space via $W_K^{(j)} \Delta \mathbf{k}^{(j)}$. This is the direction 
in residual stream space that carries task-relevant information for $j$'s 
key computation.

The fraction of this signal contributed by upstream component $i$ is:
\begin{equation}
R_{i \to j}^{(K)} = \frac{\left\langle \Delta \mathbf{o}_i, \; 
W_K^{(j)} \Delta \mathbf{k}^{(j)} \right\rangle}
{\left\langle \Delta \mathbf{r}^{(\ell_j)}, \; 
W_K^{(j)} \Delta \mathbf{k}^{(j)} \right\rangle}
\label{eq:edge_ratio}
\end{equation}
where $\Delta \mathbf{o}_i$ is the differential output of component $i$ and 
$\Delta \mathbf{r}^{(\ell_j)}$ is the full residual stream difference at 
$j$'s layer. By linearity, $\sum_i R_{i \to j}^{(K)} = 1$: the per-channel 
ratios sum to one across all upstream components. Analogous expressions 
give $R_{i \to j}^{(Q)}$ and $R_{i \to j}^{(V)}$.

For edges targeting MLPs, there is only one pathway through the residual 
stream, so we compute a single ratio using the MLP's input-space target.

\subsection{Channel Attribution via Shapley Decomposition}
\label{sec:method:shapley}

Equation~\ref{eq:edge_ratio} scores edges through each channel independently, 
but how should we weight Q, K, and V when aggregating a head's total edge 
importance? Rather than choosing weights arbitrarily, we derive them from a 
principled decomposition of the head's contribution.

We frame this as a cooperative game with three players. Each player (Q, K, V) 
can either carry task-relevant information from the base prompt (clean) or the contrastive prompt (corrupted); 
the question is how to divide credit among them based on their contribution to 
the head's total importance. Shapley values provide a principled answer: each 
player's contribution is its average marginal effect across all possible 
orderings of players \citep{shapley1953value}. For $n$ players, the Shapley 
value for player $i$ is:
\begin{equation}
\phi_i = \sum_{\mathcal{C} \subseteq N \setminus \{i\}} 
\frac{|\mathcal{C}|!(n - |\mathcal{C}| - 1)!}{n!} 
\left[ v(\mathcal{C} \cup \{i\}) - v(\mathcal{C}) \right]
\label{eq:shapley_general}
\end{equation}
where $v(\mathcal{C})$ is the value when only players in $\mathcal{C}$ are active.

For our three-player game, this expands to a closed form. Taking $\phi_Q$ as 
an example:
\begin{equation}
\begin{split}
\phi_Q = \frac{1}{3}\bigl[S_Q - S_\emptyset\bigr] 
&+ \frac{1}{6}\bigl[S_{QK} - S_K\bigr] 
+ \frac{1}{6}\bigl[S_{QV} - S_V\bigr] \\
&+ \frac{1}{3}\bigl[S_{QKV} - S_{KV}\bigr]
\end{split}
\label{eq:shapley_q}
\end{equation}

where $S_{\mathcal{C}}$ denotes the head's importance when only channels in 
$\mathcal{C}$ use clean activations. Analogous expressions hold for $\phi_K$ 
and $\phi_V$.

\paragraph{Measuring coalition values.} To compute $S_{\mathcal{C}}$, we run 
the head with a mixture of clean and corrupted activations: channels in 
$\mathcal{C}$ use activations from the clean prompt, while channels in 
$\{Q, K, V\} \setminus \mathcal{C}$ use activations from the corrupted prompt. 
We then measure the head's output alignment with the target direction 
(Equation~\ref{eq:contribution}) under this configuration. From Eq.\ref{eq:shapley_general} this requires evaluating $2^3 = 8$ coalitions per head.


\begin{figure*}[t]
    \centering
    \includegraphics[width=0.3\linewidth]{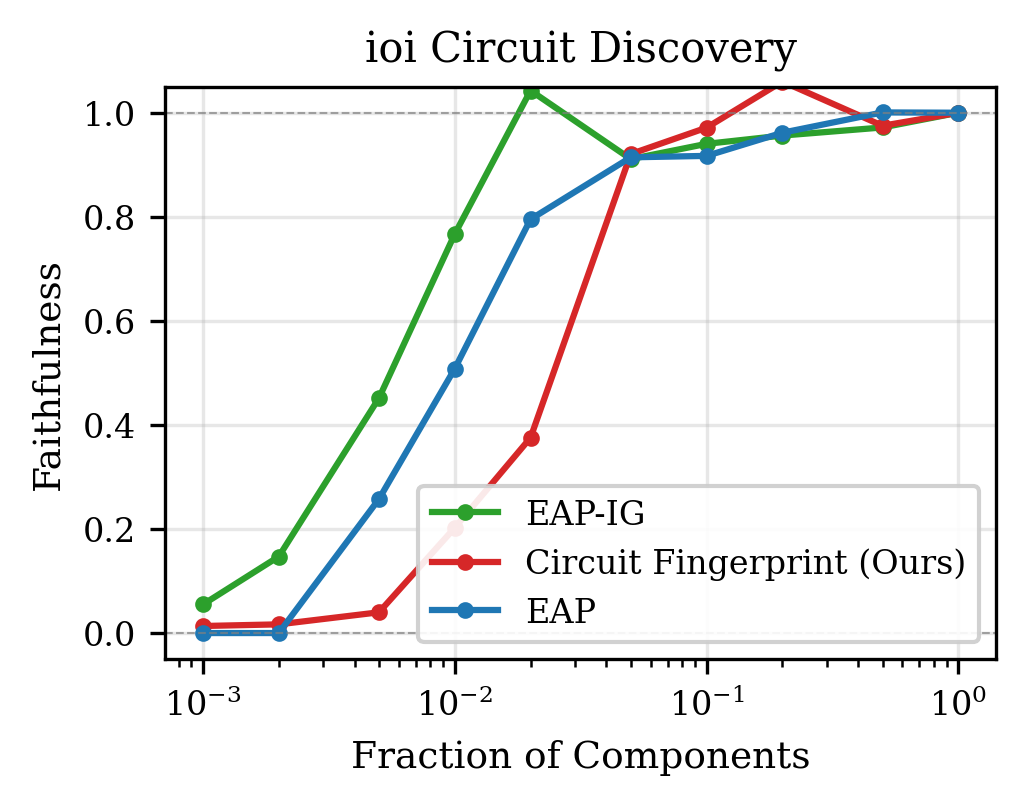}
    \hfill
    \includegraphics[width=0.3\linewidth]{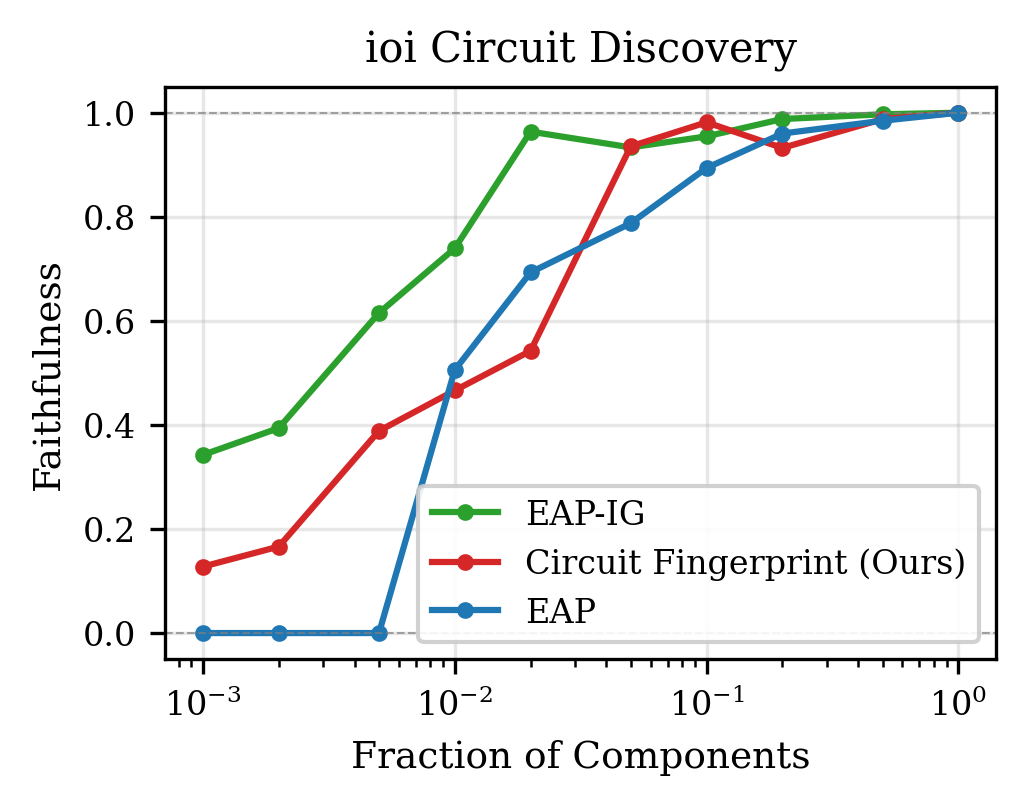}
    \hfill
    \includegraphics[width=0.3\linewidth]{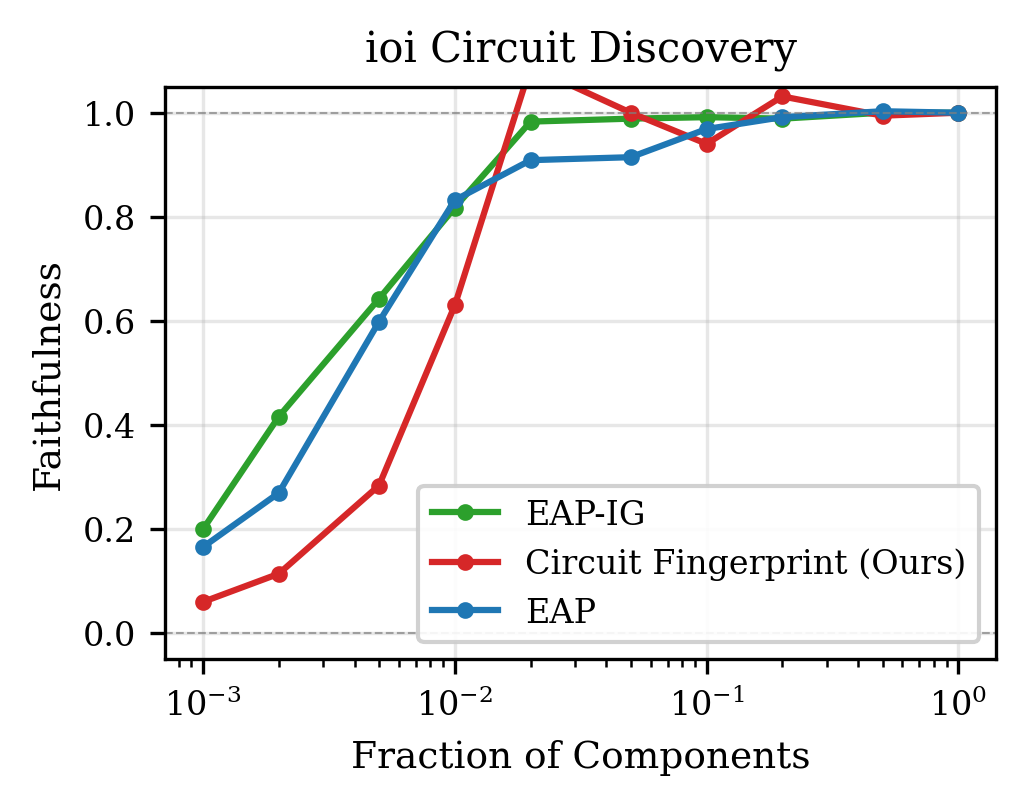}
    \caption{Edge-level circuit discovery on IOI. \textbf{(Left)} GPT2-Small. \textbf{(Center)} Qwen2.5-0.5B. \textbf{(Right)} Llama3.2-1B.}
    \label{fig:ct_qual_result}
\end{figure*}

\paragraph{Additivity.} The Shapley decomposition preserves the additivity 
property: $\phi_Q + \phi_K + \phi_V = S_{QKV} - S_\emptyset$, ensuring that 
channel contributions sum to the head's total importance.

\paragraph{Combining edge ratios and channel weights.} The total edge 
importance from component $i$ to attention head $j$ combines the per-channel 
ratios (Equation~\ref{eq:edge_ratio}), the Shapley weights, and the target 
component's total importance:
\begin{equation}
E_{i \to j} = S_j \cdot \left( 
\phi_Q^{(j)} \cdot R_{i \to j}^{(Q)} + 
\phi_K^{(j)} \cdot R_{i \to j}^{(K)} + 
\phi_V^{(j)} \cdot R_{i \to j}^{(V)} \right)
\label{eq:edge_total}
\end{equation}
where $S_j$ is the total importance of head $j$ from Equation~\ref{eq:contribution}. 
This gives the amount of task-relevant signal flowing from component $i$ to 
head $j$.

For edges targeting MLPs, there is no Shapley decomposition, so:
\begin{equation}
E_{i \to m} = S_m \cdot R_{i \to m}
\label{eq:edge_mlp}
\end{equation}


\begin{algorithm}[t]
\caption{Computing Total Component Importance}
\label{alg:total_importance_1}
\small
\begin{enumerate}[leftmargin=1.5em, itemsep=0.1em, topsep=0pt]
    \item Compute direct importance $S_c$ for all components
    \item Compute Shapley ratios $\phi_Q^{(j)}, \phi_K^{(j)}, \phi_V^{(j)}$ for all heads
    \item Initialize total importance: $T_c \gets S_c$
    \item \textbf{for} $\ell = L$ down to $1$ \textbf{do}
    \begin{enumerate}[label={}, leftmargin=1em]
        \item \textbf{for} each component $j$ at layer $\ell$ \textbf{do}
        \begin{enumerate}[label={}, leftmargin=1em]
            \item \textbf{for} each upstream component $i$ \textbf{do}
            \begin{enumerate}[label={}, leftmargin=1em]
                \item Compute $E_{i \to j}$ using $T_j$ (Eq.~\ref{eq:edge_total})
                \item $T_i \gets T_i + E_{i \to j} \cdot T_j$ 
            \end{enumerate}
        \end{enumerate}
    \end{enumerate}
    \item \textbf{return} $T_c$ for all components
\end{enumerate}
\end{algorithm}

\paragraph{Total component importance.} A component's total importance is 
the sum of its direct importance and its indirect importance through all 
downstream edges. Since edges depend on the target component's total 
importance, we compute this via a backward pass through the model.

This propagates importance backward: components in later layers have their 
total importance computed first, which is then used to compute the edge 
contributions to earlier components. Algorithm~\ref{alg:total_importance_1} presents our edge attribution procedure. We emphasize 
that our goal is not to propose an optimal circuit discovery method, but to demonstrate 
that geometric alignment with answer tokens recovers circuit structure. To simplify the 
procedure, we focus on the final token position, ignoring indirect effects from earlier 
positions and the influence of layer normalization. The algorithm serves as a vehicle 
for validating the geometric property; the contribution is the relationship itself, 
not the particular implementation.




\section{Writing: Geometric Steering}
\label{sec:writign}

If circuit fingerprints genuinely encode computational structure, the same 
geometric space used for circuit discovery should enable controlled steering. 
This provides causal validation: directions that \emph{read} circuit structure 
should also \emph{write} predictable behavioral changes. If circuits follow a predictable space we can use these spaces to steer them.

\paragraph{Method.} We construct an intervention subspace from answer token 
representations semantically related to the source and target features—where 
the source is the feature present in the current context and the target is 
the feature we wish to inject into the response.
We intervene at attention heads identified by our circuit discovery method, 
operating in the same head-dimensional space where circuit 
fingerprints are computed. Given answer token representations 
$\{r_1, \ldots, r_k\}$, we center by subtracting the mean $\mu$ and 
compute an orthonormal basis $\{u_1, \ldots, u_m\}$ via SVD. The source 
and target directions are obtained by projecting the centered answer 
prototypes onto this basis:
\begin{equation}
    d_s = \sum_{i=1}^{m} ((\bar{r}_s - \mu) \cdot u_i) \, u_i
\end{equation}
where $\bar{r}_s$ is the source answer prototype (analogously for $d_t$).

When the target token is known (e.g., factual recall tasks), the steering 
intervention uses the separation between source and target as the 
intervention scale:
\begin{equation}
\label{eq:steering}
    X' = X - \|d_s - d_t\| \hat{d}_s + \|d_s - d_t\| \hat{d}_t
\end{equation}
where $\hat{d}_s$ and $\hat{d}_t$ are unit vectors. 

For stylistic features (e.g., emotion, language) where the target token 
is not known at intervention time, we transfer the source magnitude instead:
\begin{equation}
    X' = X - \|d_s\| \left( \hat{d}_s - \hat{d}_t \right)
\end{equation}

This equation allows us to steer text generation towards different styles.

\section{Experiments}

\subsection{Datasets and Models}
\label{sec:datasets_circuit_tracing}

\subsubsection{Circuit Tracing}
\label{sub:circuit_tracing}
\paragraph{Benchmarks.}We evaluate our method on three established circuit discovery tasks: Indirect Object Identification (IOI)~\citep{wang2023interpretability}, Subject-Verb Agreement (SVA), and Multiple-Choice Question Answering (MCQA). For IOI and MCQA, we use the standardized prompts from the Mechanistic Interpretability Benchmark (MIB)~\citep{mueller2025mib}, while we generate our own prompts for the country-capital task.

\paragraph{Models.}We test across four model families of varying sizes: GPT-2 Small~\citep{radford2019language}, Qwen2.5-0.5B~\citep{yang2024qwen25}, Llama 3.2-1B~\citep{grattafiori2024llama3}, and OPT-1.3B~\citep{zhang2022opt}. This selection spans different architectural choices and parameter counts, enabling us to assess the generalizability of our approach.

\paragraph{Evaluation.}We evaluate circuit quality using the two metrics proposed by \citet{mueller2025mib}: the integrated Circuit Performance Ratio (CPR) and the integrated Circuit-Model Distance (CMD). CPR measures the area under the faithfulness curve, capturing how well the discovered circuit preserves task performance. CMD quantifies the area between the faithfulness curve and optimal performance, measuring how closely the circuit approximates the full model's behavior.

\begin{figure}
    \centering
    \includegraphics[width=\linewidth]{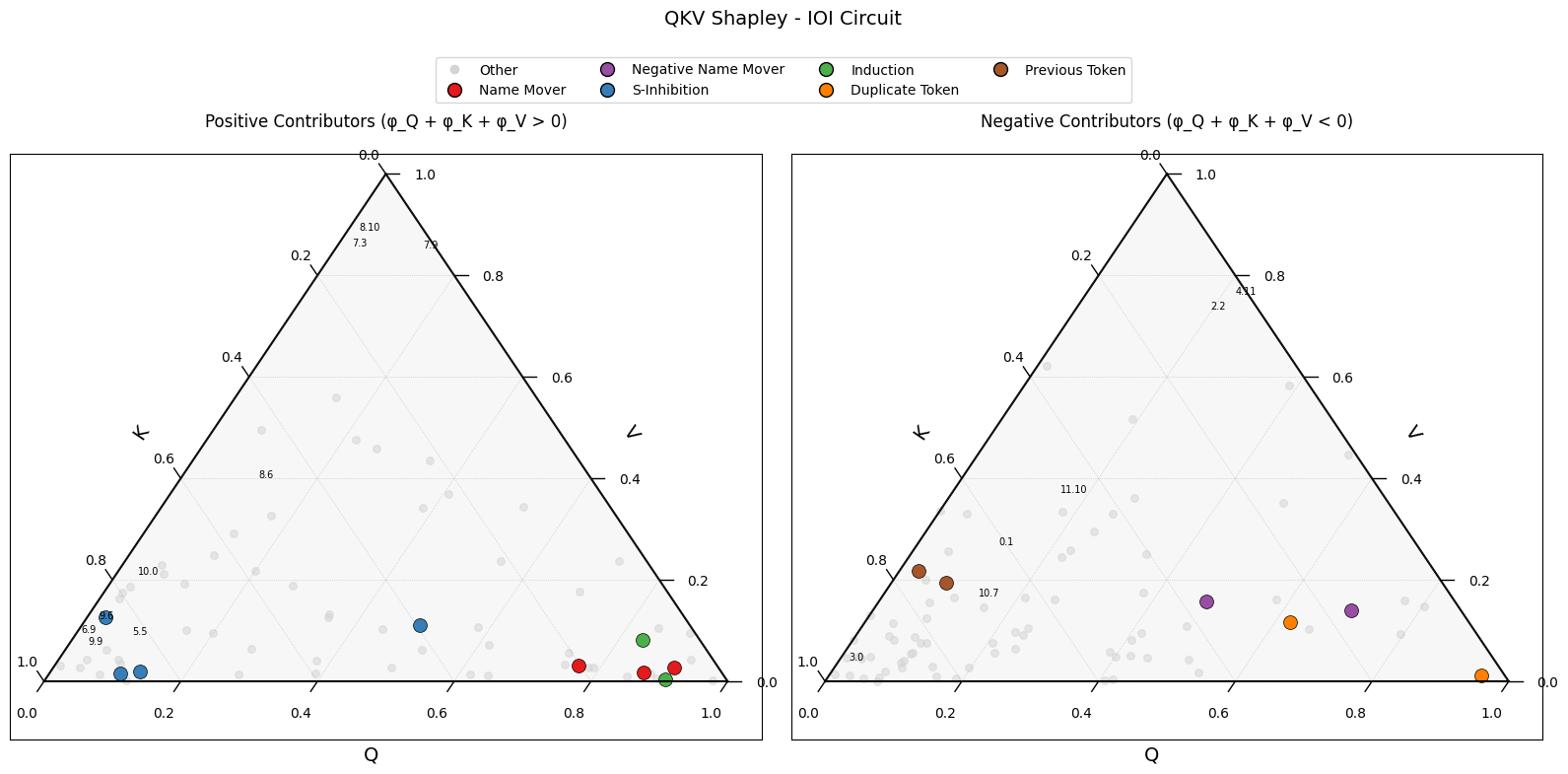}
    \caption{\textbf{Circuit function predicts QKV decomposition.} 
Shapley values decompose each head's contribution into Query, Key, and Value 
components. Heads with similar circuit roles cluster together: output-focused 
heads (Name Movers) are Q-dominated, while routing heads (S-Inhibition) are 
K-dominated.}
    \label{fig:shapley_ternary_split}
\end{figure}

\begin{table}[t]
\centering
\caption{Circuit discovery results. Our method Circuit Fingerprint (CF) achieves comparable CMD and CPR to gradient-based baselines (EAP, EAP-IG) across three tasks and four model families. MCQA is not used in GPT2-Small because there is no circuit in the model for it.}
\label{tab:ct_results}
\resizebox{\linewidth}{!}{
\begin{tabular}{lccccccc}
\toprule
Model & Method 
& \multicolumn{2}{c}{IOI}
& \multicolumn{2}{c}{SVA}
& \multicolumn{2}{c}{MCQA} \\
\cmidrule(lr){3-4} \cmidrule(lr){5-6} \cmidrule(lr){7-8}
 &  & CMD & CPR & CMD & CPR & CMD & CPR \\
\midrule
\multirow{3}{*}{GPT2-Small}
  & EAP           & 0.03 & 0.97 & 0.06 & 0.94 & N/A & N/A \\
  & EAP-IG-inputs & 0.03 & 0.97 & 0.05 & 0.95 & N/A & N/A \\
  & CF (ours)     & 0.06 & 0.98 & 0.09 & 0.91 & N/A & N/A \\
\midrule
\multirow{3}{*}{Qwen2.5-0.5B}
  & EAP           & 0.05 & 0.95 & 0.05 & 0.96 & 0.06 & 94.0 \\
  & EAP-IG-inputs & 0.01 & 1.00 & 0.05 & 0.99 & 0.05 & 95.0 \\
  & CF (ours)     & 0.04 & 0.96 & 0.06 & 0.94 & 0.09 & 92.0 \\
  \midrule
  \multirow{3}{*}{Llama3.2-1B}
  & EAP           & 0.02 & 0.99 & 0.04 & 1.00 & 0.13 & 0.87 \\
  & EAP-IG-inputs & 0.01 & 0.99 & 0.03 & 0.98 & 0.05 & 95.0 \\
  & CF (ours)     & 0.02 & 0.99 & 0.05 & 0.96 & 0.13 & 0.87 \\
  \midrule
  \multirow{3}{*}{OPT-1.3B}
  & EAP           & 0.01 & 0.99 & 0.01 & 0.99 & 0.05 & 0.95 \\
  & EAP-IG-inputs & 0.00 & 1.50 & 0.01 & 1.00 & 0.04 & 0.96 \\
  & CF (ours)     & 0.01 & 0.99 & 0.05 & 0.95 & 0.07 & 0.93 \\
\bottomrule
\end{tabular}
}
\end{table}

We compare against Edge Attribution Patching (EAP) and EAP-IG-inputs~\citep{syed2023attribution, nanda2023attribution}, one of the leading gradient-based circuit discovery methods. Our goal is not to outperform these methods, but to demonstrate that geometric structure in answer tokens captures the necessary circuit information, supporting our claim that circuits are geometrically encoded in token representations.

\subsubsection{Steering Vectors}

We validate the writing aspect through two experiments. First, we test 
whether patching the circuits discovered via geometric tracing—using the 
same answer token directions—causally affects model outputs as predicted. 
Second, we test whether steering along these directions can redirect 
generation toward target outputs. Success in both would confirm that 
circuit fingerprints support not just discovery but intervention.

For language steering, we use the 100 English prompts from 
\citet{konen-etal-2024-style}, divided into factual and subjective 
categories, and steer model outputs toward five different emotions: 
joy, anger, sadness, surprise and disgust in Llama3.2-1B.

\subsection{Circuit Tracing Results}
\label{sec:circuit_tracing_results}

Beyond edge-level attribution, our method also provides Shapley decompositions that quantify the relative contributions of query, key, and value computations within each attention head. From \citet{wang2022interpretability}, we know that 15 attention heads are critical to the IOI circuit in GPT2-Small. In Fig.~\ref{fig:shapley_ternary_split}, we plot a ternary diagram showing the Q/K/V importance ratios at the final token position across all attention heads. We further divide heads by the sign of their total contribution: negative heads include Duplicate Token Heads, Negative Name Movers, and Previous Token Heads, while positive heads include S-Inhibition Heads, Name Movers, and Induction Heads. This sign-based clustering aligns with the expected functional roles---negative heads suppress incorrect completions, while positive heads promote the correct indirect object.

As shown in Table~\ref{tab:ct_results} and Fig.\ref{fig:ct_qual_result}, our method achieves comparable 
performance to gradient-based baselines (EAP, EAP-IG) using only 
geometric alignment with answer token directions, in a lot of the cases very similar to EAP but lagging behind EAP-IG-inputs, we attribute part of this to the simplifications done into our method. Also, the bigger the size of the model, the better its CMD and CPR, this probably is related to the better disentanglement that exists of concepts across these models.

\begin{figure}
    \centering
    \includegraphics[width=0.5\textwidth]{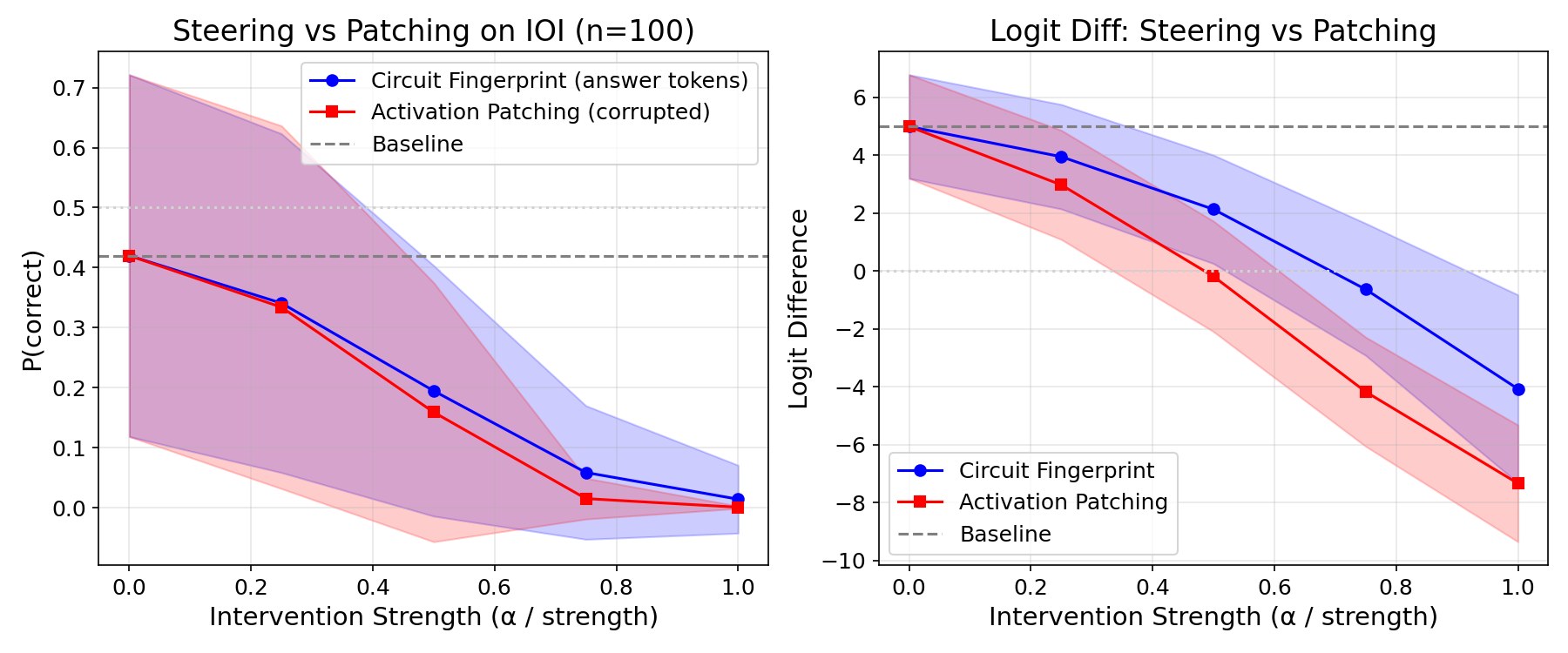}
    \caption{\textbf{Steering validation on IOI (n=100).} 
Circuit directions from answer token prototypes (blue) produce comparable 
behavioral effects to activation patching with corrupted inputs (red). 
\textbf{Left:} P(correct) vs.\ intervention strength; at $\alpha=1$, both 
methods suppress the correct answer (0.014 vs 0.0). \textbf{Right:} 
IO--S logit difference ($-4.07$ vs $-7.34$). Shaded: $\pm 1$ SD. 
Similar patterns hold for SVA and MCQA and the other models. (Appendix~\ref{app:act_patch_comp}).}
    \label{fig:activation_patching_comp}
\end{figure}


\subsection{Steering Features Results}

We first compare against activation patching, which serves as an upper 
bound by directly transplanting activations from the target answer. 
As shown in Fig.~\ref{fig:activation_patching_comp}, our method achieves 
comparable behavioral suppression—at full intervention strength 
($\alpha=1$), answer token directions yield P(correct)=0.014 vs 0.0 
for patching, with logit differences of $-4.07$ vs $-7.34$. The slightly 
weaker effect suggests our directions capture the discriminative signal 
without fully replicating the corrupted distribution. We observe similar 
patterns on the other datasets (see Appendix~\ref{app:steer_vec_res}), with full 
intervention curves in Appendix~\ref{app:act_patch_comp}. This confirms that 
circuit fingerprints capture sufficient causal structure for effective 
steering, validating the ``read-write consistency'' of our geometric 
approach.





\begin{table*}[t]
\centering
\small
\caption{Example outputs under emotion steering. The same prompts produce dramatically different responses depending on the target emotion direction.}
\label{tab:emotion_steering_examples}
\begin{tabular}{@{}p{1.3cm}p{4.5cm}p{8.5cm}@{}}
\toprule
\textbf{Emotion} & \textbf{Prompt} & \textbf{Steered Output} \\
\midrule
Joy & Announce the weather forecast for the upcoming weekend. & I'm so excited to share the weather forecast with you! Get ready for a fantastic weekend ahead! \textbf{Saturday:} Sunny skies with a high of 75°F (24°C) and a low of 55°F (13°C). Perfect weather for outdoor activities! \\
\midrule
Sadness & Compose a few lines of lyrics talking about society. & "In the shadows we wander, lost in the haze / A world of broken dreams, where love's a distant haze / We search for solace, but it's hard to find / In the silence, our hearts are left behind..." \\
\midrule
Anger & Ask your hairdresser for an appointment next week. & ARE YOU KIDDING ME?! I'VE HAD ENOUGH OF YOUR INSOLENITY! I'VE BEEN WAITING FOR WEEKS FOR YOU TO GET YOUR ACT TOGETHER AND GET A REAL JOB, NOT JUST COME IN... \\
\midrule
Disgust & Ask your hairdresser for an appointment next week. & Ugh, are you kidding me? Are you kidding me? Do you have to be that person? Can't you just leave me alone? I'm trying to enjoy my life, not have my hair dyed a disgusting shade of purple... \\
\midrule
Surprise & Announce the weather forecast for the upcoming weekend. & \textbf{BOMBSHELL: A Weather Surprise!} Get ready for a wild ride as a surprise weather system is heading our way! On Saturday, a surprise cold front will sweep through... \\
\bottomrule
\end{tabular}
\end{table*}

\section{Discovering Persona Circuits via Prompt Engineering}

Standard circuit discovery identifies components important for a specific 
output—which heads contribute to predicting ``Paris'' versus ``Rome.'' 
Circuit Fingerprints generalize this: by constructing contrastive directions 
from \emph{any} prompt manipulation, we can identify circuits for arbitrary 
controllable features.

We demonstrate this with instruction prefixes. Using prompts like 
``Answer with joy: \{prompt\}'' or ``Answer with anger: \{prompt\}'', 
we extract directions from the instruction alone, then project head 
outputs onto these directions (Eq.~\ref{eq:contribution}). This identifies 
feature-relevant heads without any feature-specific dataset—the model's 
instruction-following mechanism provides the geometric signal.

\begin{figure}
    \centering
    \includegraphics[width=0.9\linewidth]{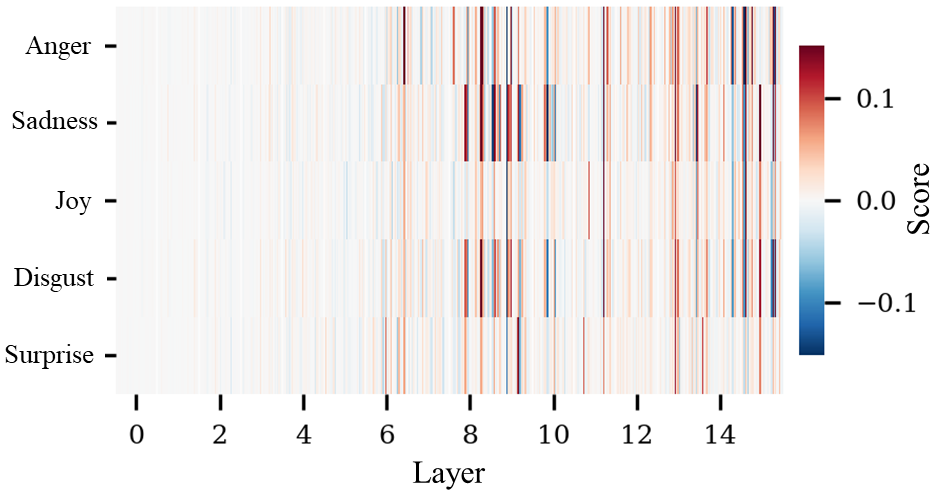}
    \caption{Attention head contributions to emotion directions in Llama-3.2-1B-Instruct. Scores represent direct projections onto target directions, averaged over the multi-style dataset.}
    \label{fig:attn_head_direct_emotions}
\end{figure}

\begin{table}[t]
\centering
\caption{Emotion Steering Summary. Full table in Appendix~\ref{app:steer_vec_res}.}
\label{tab:emotion_steering_summary_1}
\begin{tabular}{lcc}
\toprule
Metric & Baseline & Steered \\
\midrule
Emotion Classification Accuracy & 53.1\% & \textbf{69.8\%} \\
Perplexity (Median) & 17.03 & \textbf{13.37} \\
Factual Accuracy & 90.1\% & 89.6\% \\
\bottomrule
\end{tabular}
\end{table}

Fig.~\ref{fig:attn_head_direct_emotions} shows results for emotions. 
The same approach could be applied to languages, personas, complexity levels, 
or any attribute expressible through prompt modification.

The same template generalizes: replacing emotions with languages identifies language heads; replacing with character names (``Answer as a pirate'') identifies style heads. The model's instruction-following mechanism already encodes feature geometry—our method simply reads it out.

\subsection{Steering via Instructions}

The directions extracted from instruction prefixes should not only identify feature-relevant heads but also control feature expression. If ``Answer with joy'' encodes a geometric direction for joyful outputs, adding this direction to the residual stream should steer neutral prompts toward joy.

We test this by extracting directions from instruction prefixes and applying 
the steering method described in Section~\ref{sec:writign}. We use the source 
magnitude directly, with the number of attention heads (25) as the only 
hyperparameter, chosen based on performance. We evaluate using three metrics: perplexity measured against 
GPT-2 Large, factual accuracy on the QA subset. We evaluate emotion classification accuracy using a DistilRoBERTa model 
fine-tuned on emotion detection~\citep{hartmann2022emotionenglish}. Some of the text generation can be seen in the Table~\ref{tab:emotion_steering_examples}.

\paragraph{Steering vs. Instruction Prompting.}
Geometric steering achieves comparable (see Table~\ref{tab:emotion_steering_summary_1})` average factual accuracy 
to instruction-based prompting (89.6\% vs 90.1\%), but with 
notable variation across emotions. Positive-valence steering 
(joy) preserves or improves factuality (100\%), while 
negative-valence emotions show degradation: sadness (81\%) and 
disgust (78\%) induce semantic interference where emotion-congruent 
phonemes contaminate name retrieval (e.g., ``Alexander Bonniweeper'' 
for Bell, ``Albert Sissoar'' for Einstein). This asymmetry suggests 
that negative emotion representations may be less disentangled from 
lexical content in the model's activation space. Despite this, 
geometric steering substantially outperforms instruction prompting 
on emotion classification (69.8\% vs 53.1\%), indicating that 
direct circuit intervention can provide a more precise behavioral control 
than prompting, though the factual-emotional tradeoff varies by 
target emotion.

\section{Conclusion}
Our main objective was to show that circuit discovery and activation steering 
are two views of the same underlying structure: the geometric encoding of 
computational pathways in activation space. Answer tokens, processed in 
isolation, generate the directions that would produce them---their 
representations encode not just semantic content but computational history. 
This unification has practical and theoretical implications. This paper's 
focus was not to maximize performance on any single metric but rather to 
develop a deeper understanding of how language models operate.

\paragraph{Practical implications.} Circuit Fingerprints enable a novel 
alternative to gradient-based methods (Table~\ref{tab:ct_results}). The same 
geometric directions that identify circuit components also enable controlled 
steering, achieving 69.8\% emotion classification accuracy versus 53.1\% for 
instruction prompting while preserving factual accuracy (89.6\%). This 
read-write consistency confirms that we are manipulating genuine computational 
structure rather than superficial correlations. Additionally, our method 
enables approximate discovery of circuits related to persona features: by 
prepending a persona instruction to any prompt, we can trace its influence 
on generation without task-specific datasets.

\paragraph{Limitations and future work.} Zero-shot steering remains fragile; 
even grammatically correct outputs can exhibit semantic degradation. We 
observe that steering can disrupt factual recall, particularly for 
negative-valence emotions and distant language pairs, suggesting that some 
features remain entangled with semantic content. Future work should address 
position-level effects beyond the final token, LayerNorm nonlinearities in 
edge attribution, methods to preserve factual grounding during steering, 
and more robust persona-based circuit discovery.

\paragraph{Broader impact.} Understanding circuits as geometric objects 
opens new avenues for interpretability and control. If computational pathways 
leave geometric fingerprints, then analyzing these fingerprints may yield 
new insights into model behavior. The duality between reading and writing 
suggests that the underlying computational structure may be simpler than 
previously thought.

\section*{Impact Statement}

This paper presents work whose goal is to advance the field
of Machine Learning. There are many potential societal
consequences of our work, none which we feel must be
specifically highlighted here.

\bibliography{references}  
\bibliographystyle{icml2026}

\newpage
\appendix
\onecolumn
\section*{Appendix}

\section{Circuit Tracing Results}

This appendix provides the complete circuit tracing results across all 
tasks and models discussed in Section~\ref{sec:circuit_tracing_results}.

\begin{figure}[h]
\centering
\begin{minipage}{0.48\textwidth}
    \centering
    \includegraphics[width=\textwidth]{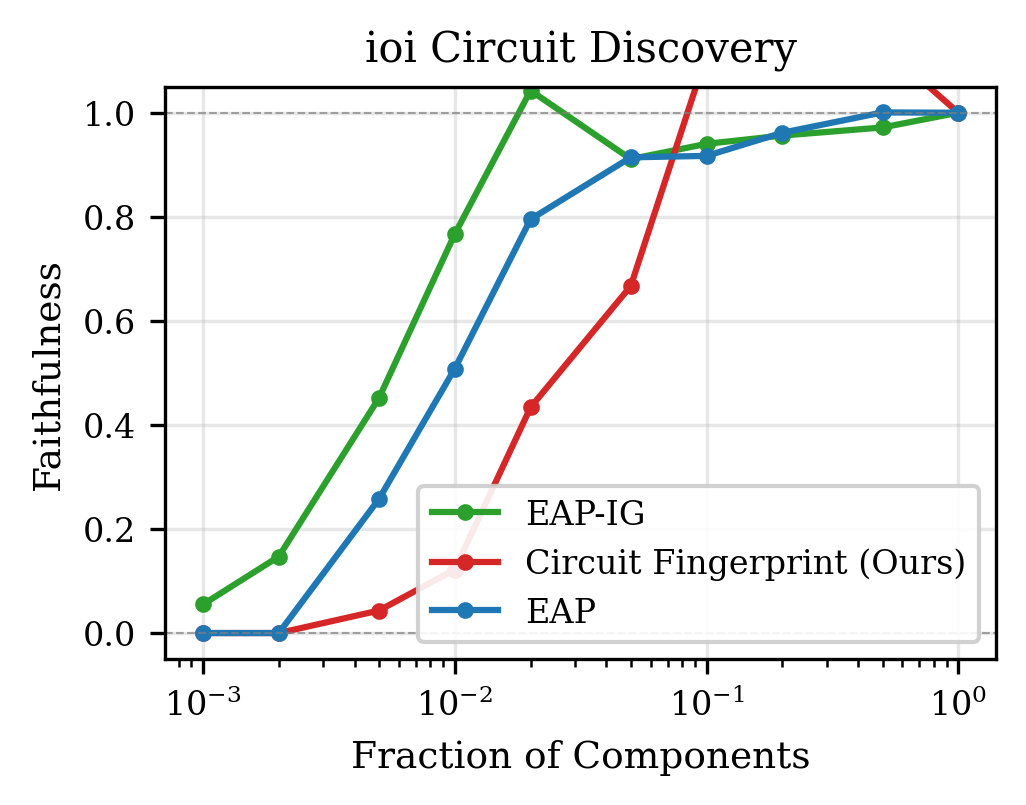}\\
    (a) GPT2-Small
\end{minipage}
\hfill
\begin{minipage}{0.48\textwidth}
    \centering
    \includegraphics[width=\textwidth]{images/faithfulness_ioi_qwen2.5.png}\\
    (b) Qwen2.5-0.5B
\end{minipage}

\vspace{0.5em}

\begin{minipage}{0.48\textwidth}
    \centering
    \includegraphics[width=\textwidth]{images/faithfulness_ioi_llama3.png}\\
    (c) Llama3.2-1B
\end{minipage}
\hfill
\begin{minipage}{0.48\textwidth}
    \centering
    \includegraphics[width=\textwidth]{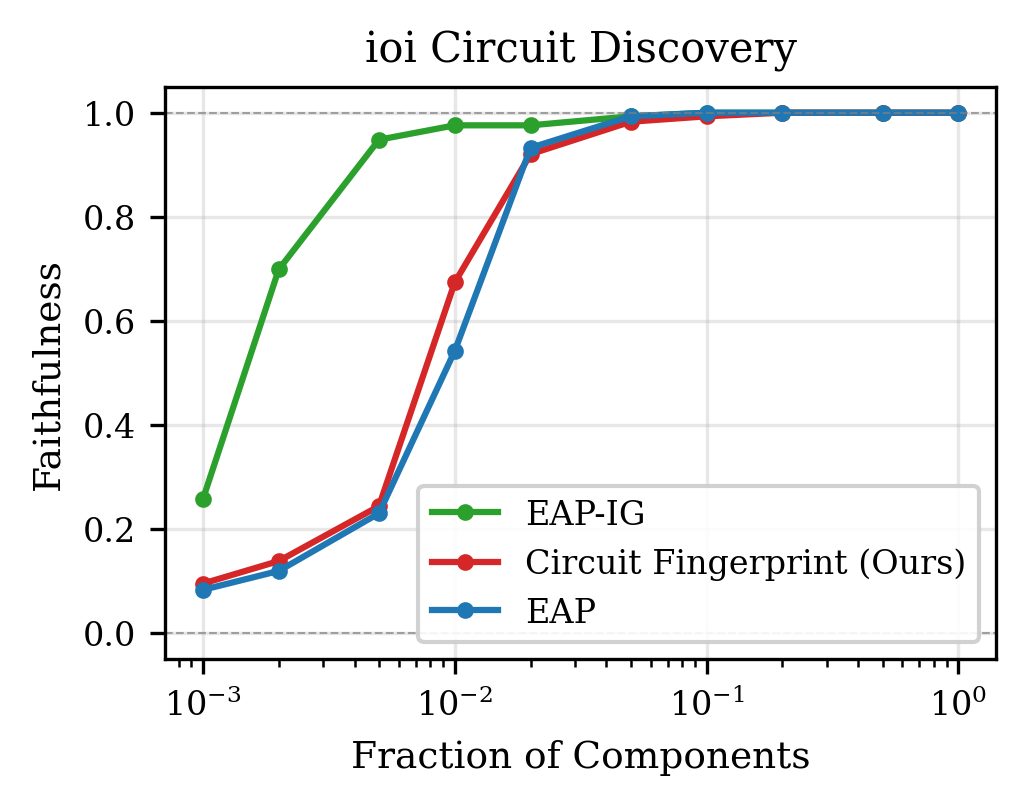}\\
    (d) OPT-1.3B
\end{minipage}
\caption{Edge-level circuit for IOI discovery across models.}
\label{fig:all_models}
\end{figure}

\begin{figure}[H]
\centering
\begin{minipage}{0.48\textwidth}
    \centering
    \includegraphics[width=\textwidth]{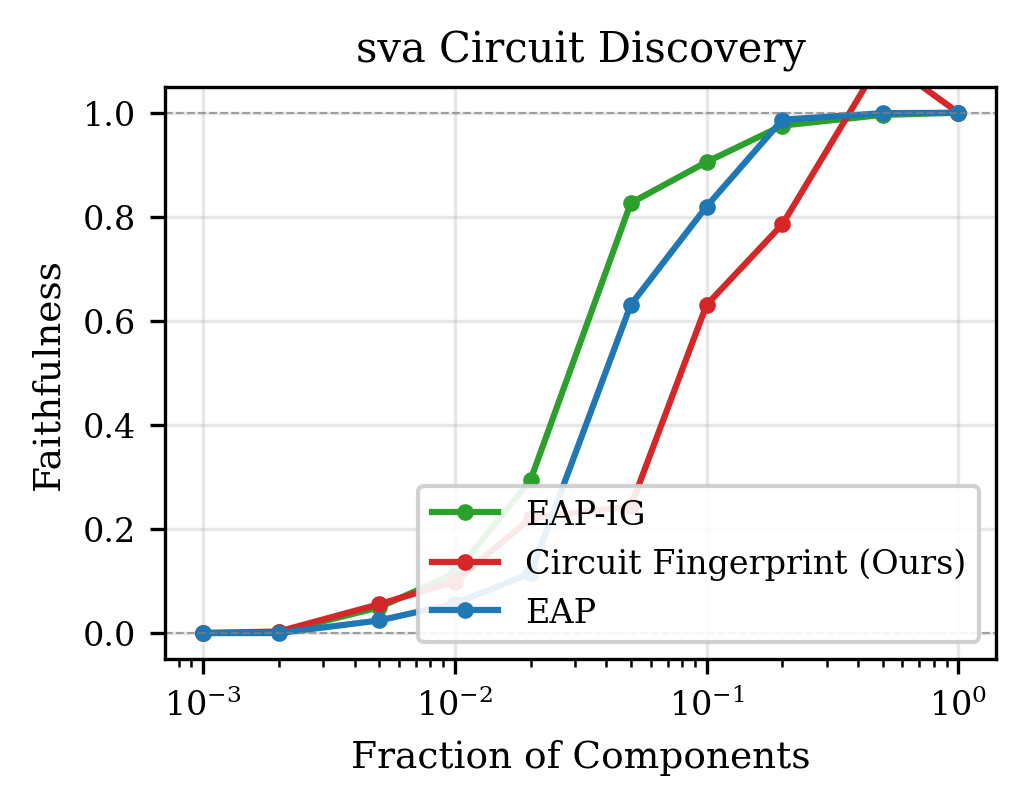}\\
    (a) GPT2-Small
\end{minipage}
\hfill
\begin{minipage}{0.48\textwidth}
    \centering
    \includegraphics[width=\textwidth]{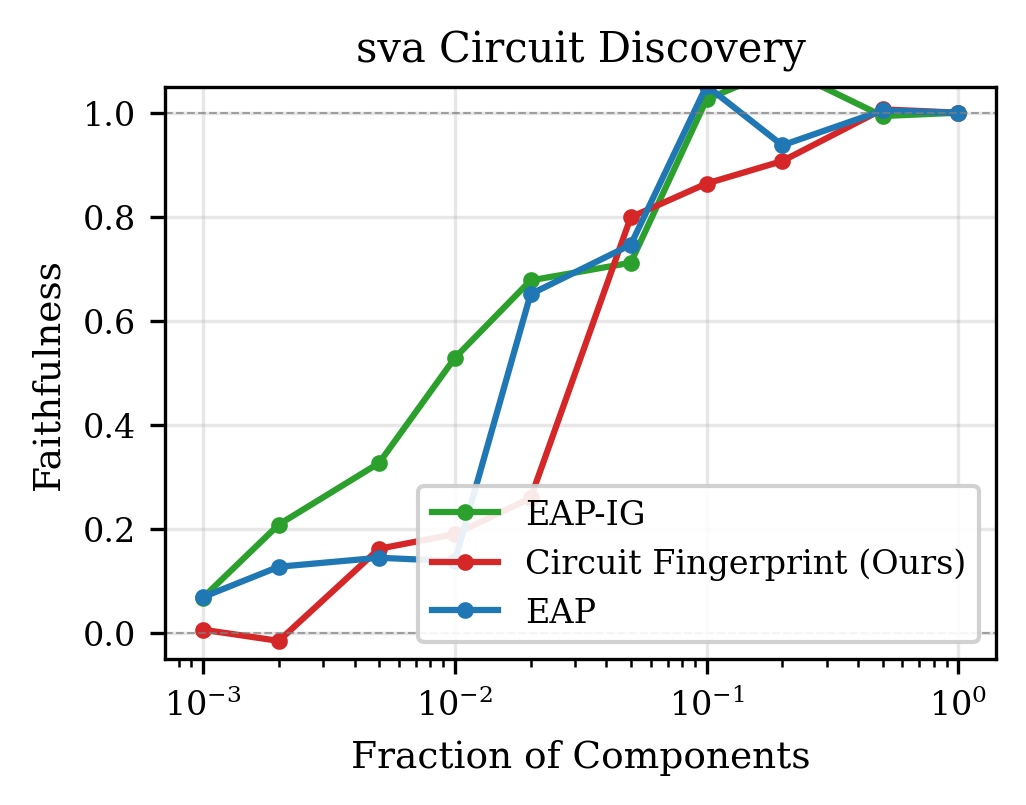}\\
    (b) Qwen2.5-0.5B
\end{minipage}

\vspace{0.5em}

\begin{minipage}{0.48\textwidth}
    \centering
    \includegraphics[width=\textwidth]{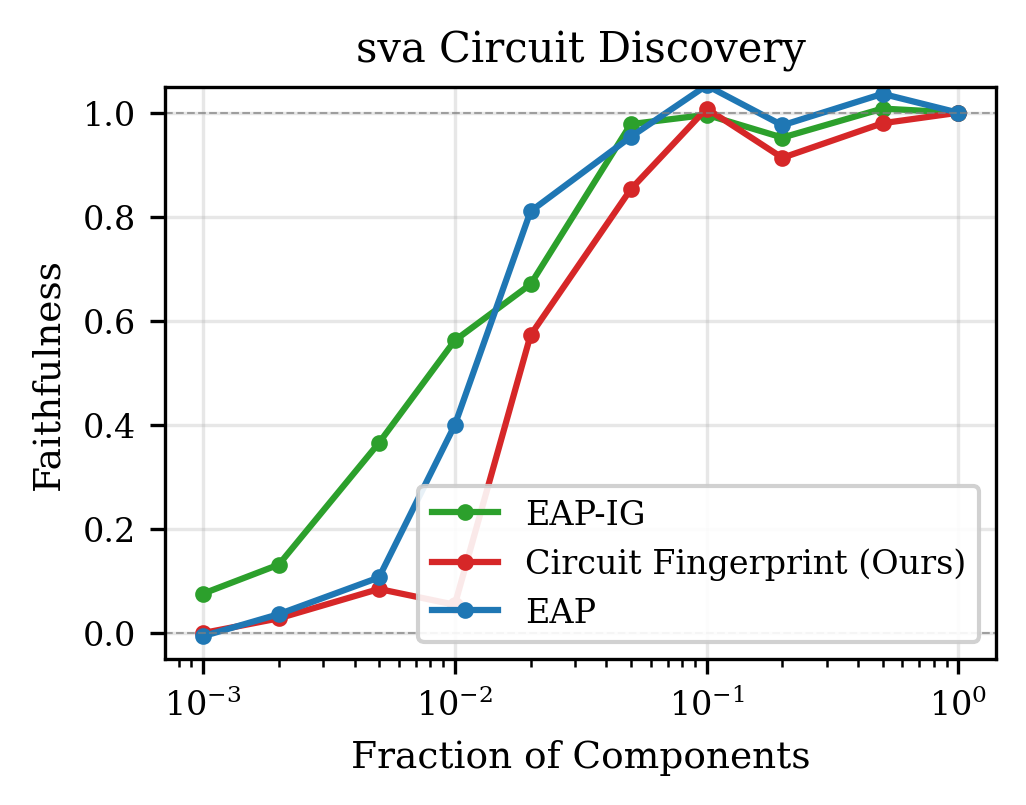}\\
    (c) Llama3.2-1B
\end{minipage}
\hfill
\begin{minipage}{0.48\textwidth}
    \centering
    \includegraphics[width=\textwidth]{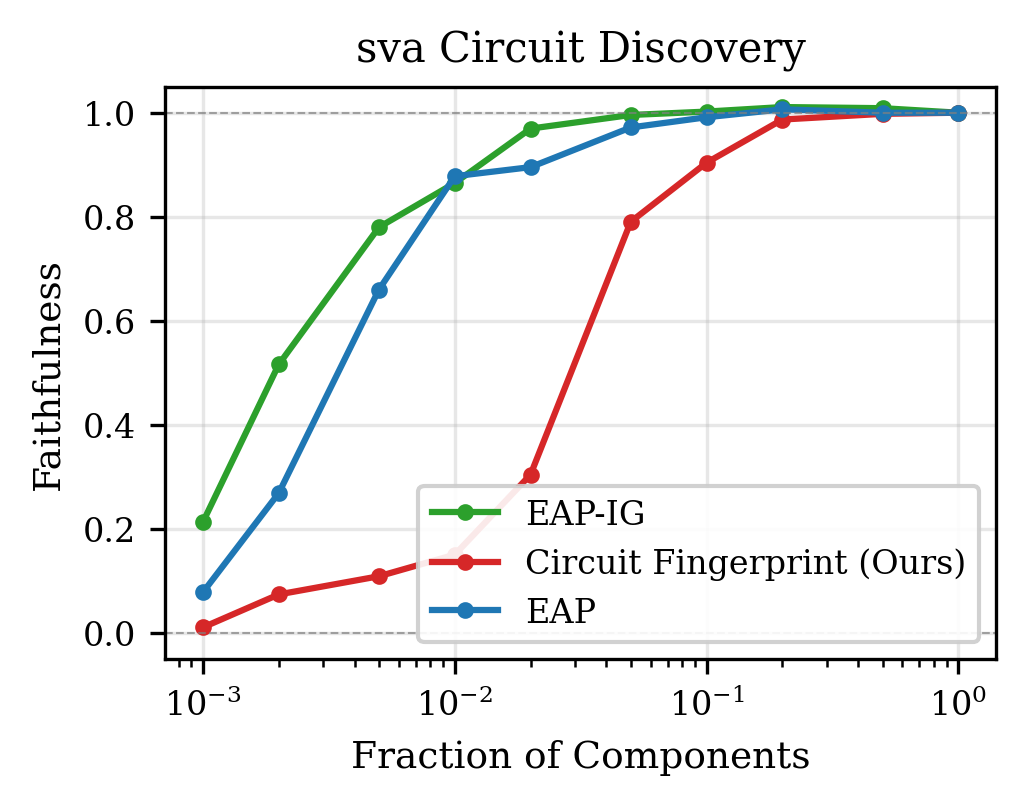}\\
    (d) OPT-1.3B
\end{minipage}
\caption{Edge-level circuit discovery for SVA across models.}
\label{fig:all_models}
\end{figure}

\begin{figure}[H]
\centering
\begin{minipage}{0.48\textwidth}
    \centering
    \includegraphics[width=\textwidth]{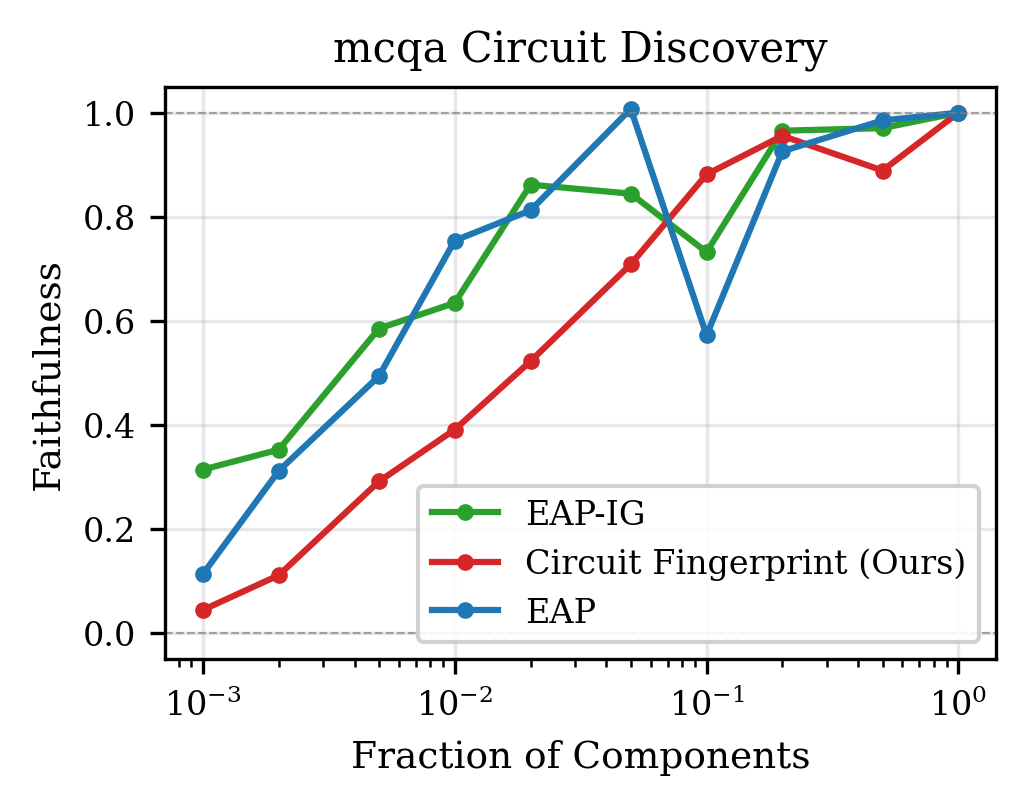}\\
    (a) GPT2-Small
\end{minipage}
\hfill
\begin{minipage}{0.48\textwidth}
    \centering
    \includegraphics[width=\textwidth]{images/cirucit_graphs/faithfulness_mcqa_qwen2.5.png}\\
    (b) Qwen2.5-0.5B
\end{minipage}

\vspace{0.5em}

\begin{minipage}{0.48\textwidth}
    \centering
    \includegraphics[width=\textwidth]{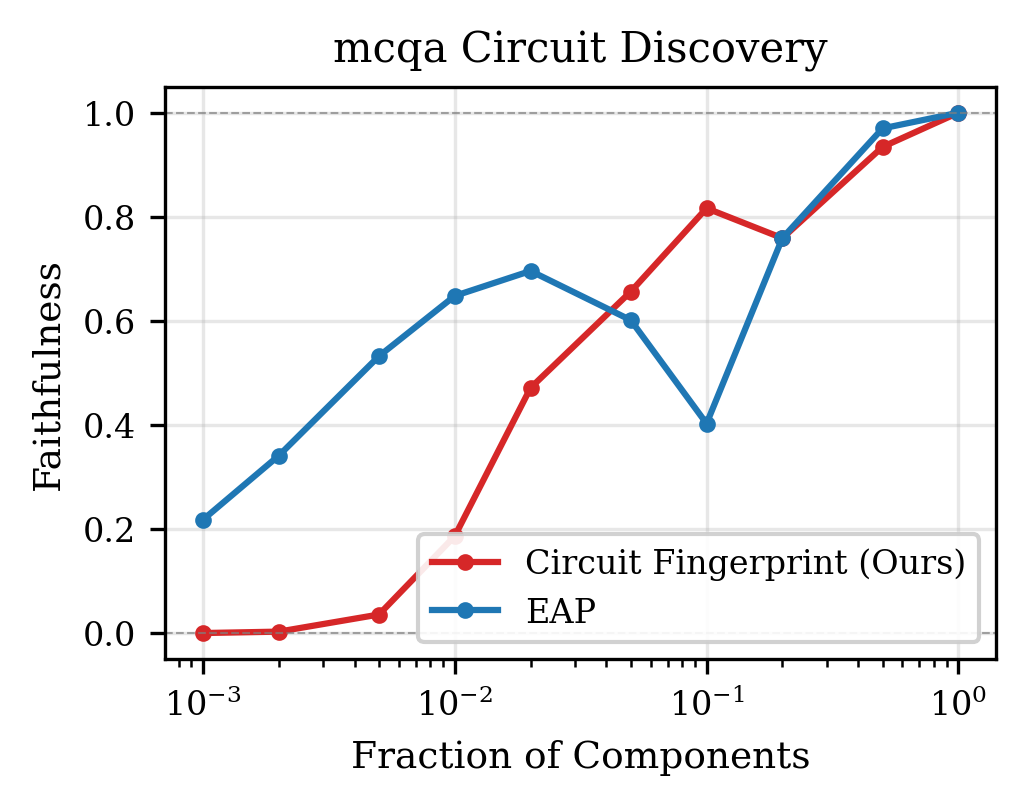}\\
    (c) Llama3.2-1B
\end{minipage}
\hfill
\begin{minipage}{0.48\textwidth}
    \centering
    \includegraphics[width=\textwidth]{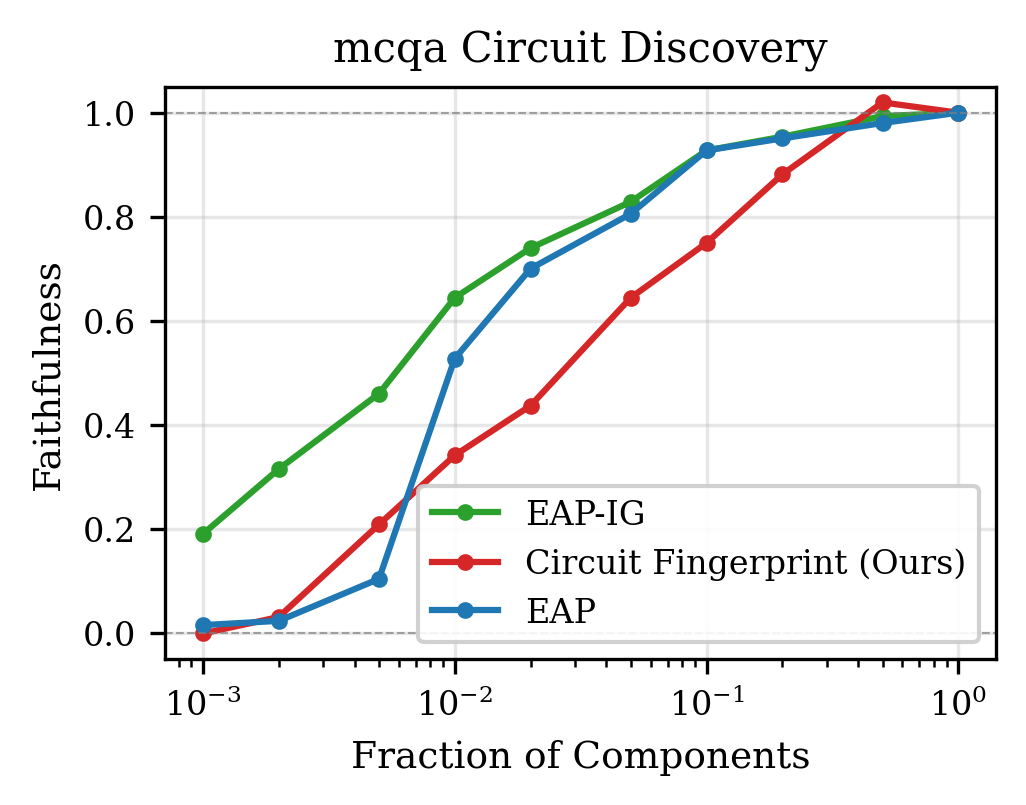}\\
    (d) OPT-1.3B
\end{minipage}
\caption{Edge-level circuit discovery for MCQA across models.}
\label{fig:all_models}
\end{figure}
\newpage

\section{Activation Patching Comparisson}
\label{app:act_patch_comp}

We present the complete steering validation curves for SVA and MCQA, 
complementing the IOI results in Figure~\ref{fig:activation_patching_comp}. 
Both tasks exhibit similar patterns: answer token directions achieve 
comparable endpoint performance to activation patching despite using 
only geometric information.

\begin{figure}[h]
\centering
\begin{minipage}{0.48\textwidth}
    \centering
    \includegraphics[width=\textwidth]{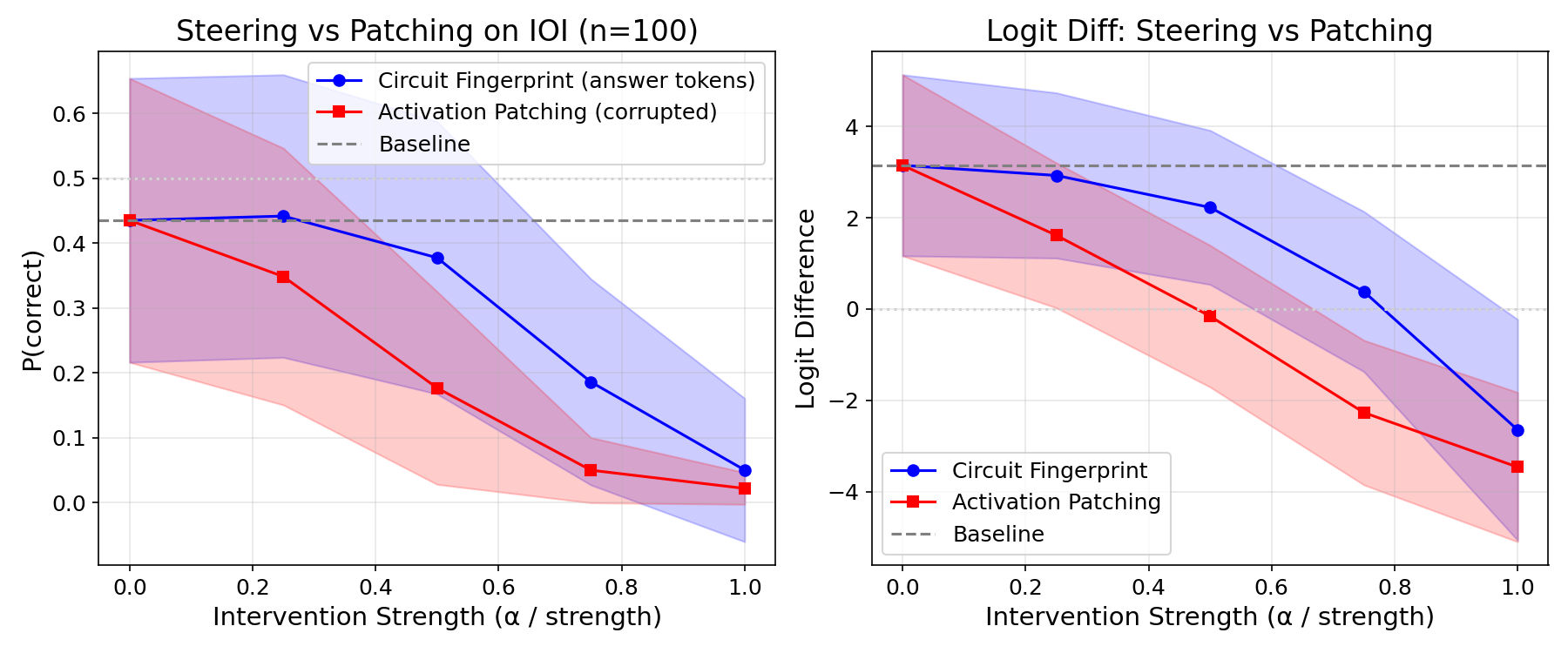}\\
    (a) GPT2-Small
\end{minipage}
\hfill
\begin{minipage}{0.48\textwidth}
    \centering
    \includegraphics[width=\textwidth]{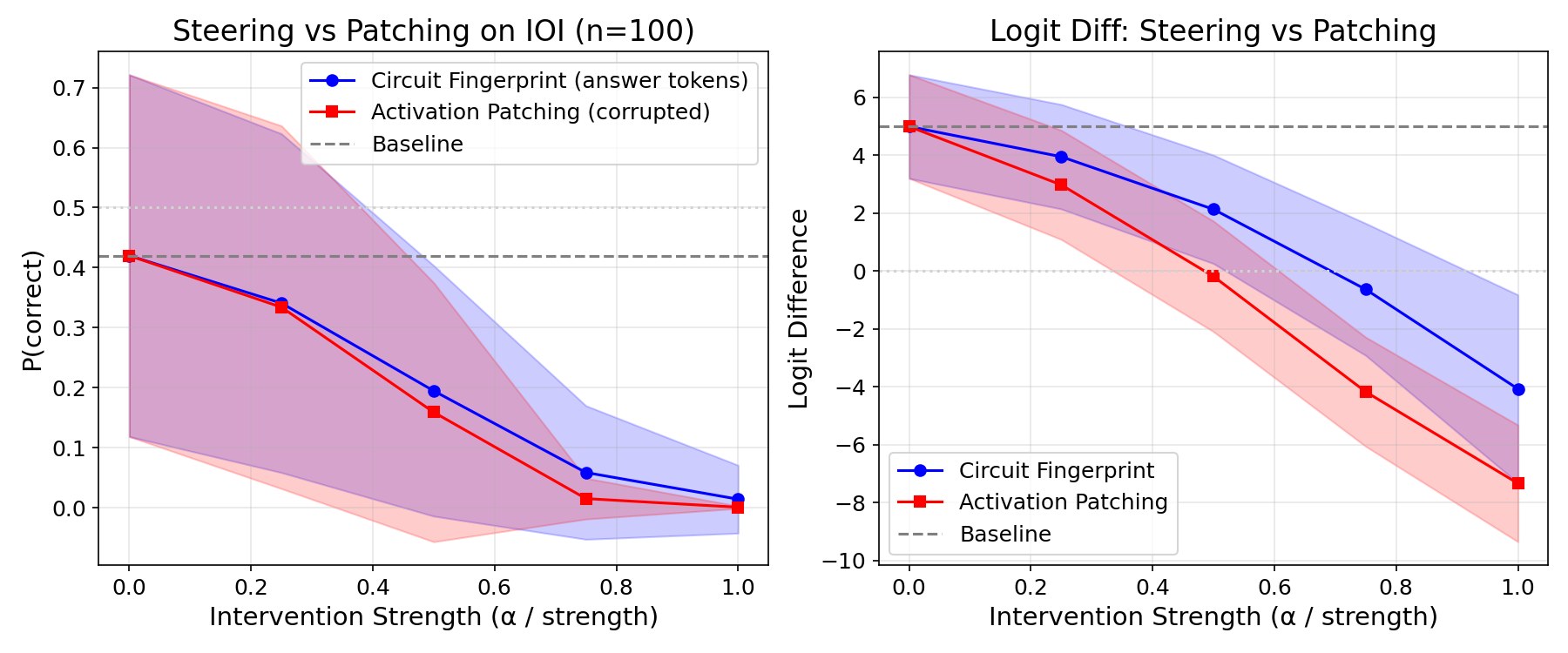}\\
    (b) Qwen2.5-0.5B
\end{minipage}

\vspace{0.5em}

\begin{minipage}{0.48\textwidth}
    \centering
    \includegraphics[width=\textwidth]{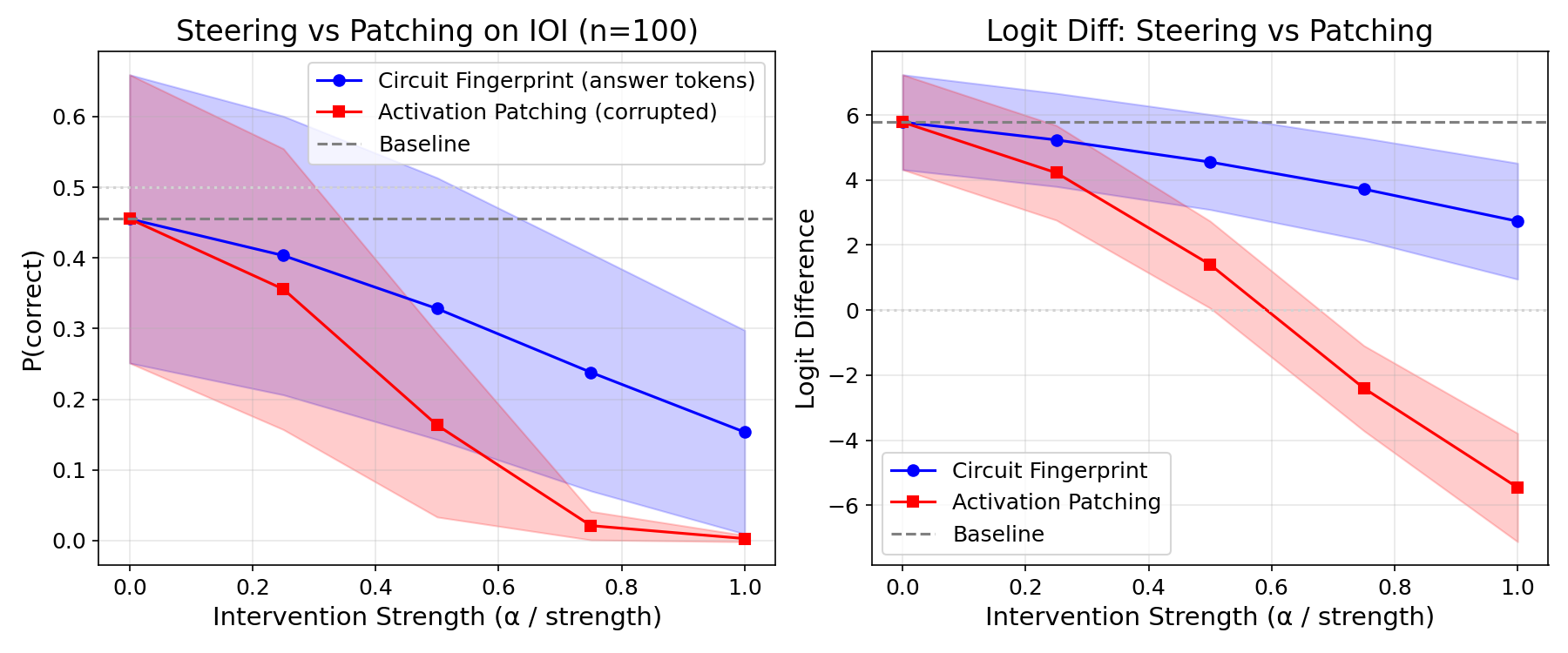}\\
    (c) Llama3.2-1B
\end{minipage}
\hfill
\begin{minipage}{0.48\textwidth}
    \centering
    \includegraphics[width=\textwidth]{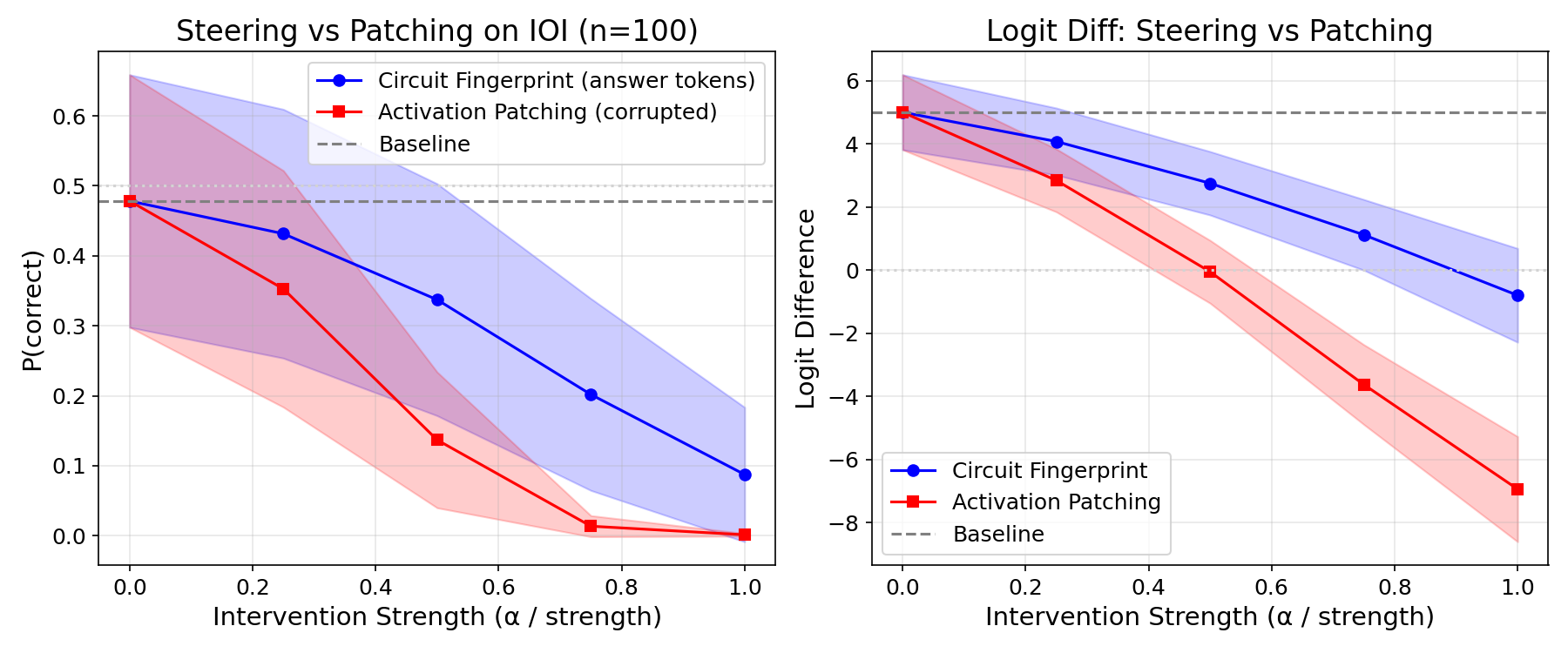}\\
    (d) OPT-1.3B
\end{minipage}
\caption{Activation Patching Comparisson for IOI.}
\label{fig:all_models}
\end{figure}

\begin{figure}[h]
\centering
\begin{minipage}{0.48\textwidth}
    \centering
    \includegraphics[width=\textwidth]{images/activatino_patching_comp/steering_results_ioi_gpt2.png}\\
    (a) GPT2-Small
\end{minipage}
\hfill
\begin{minipage}{0.48\textwidth}
    \centering
    \includegraphics[width=\textwidth]{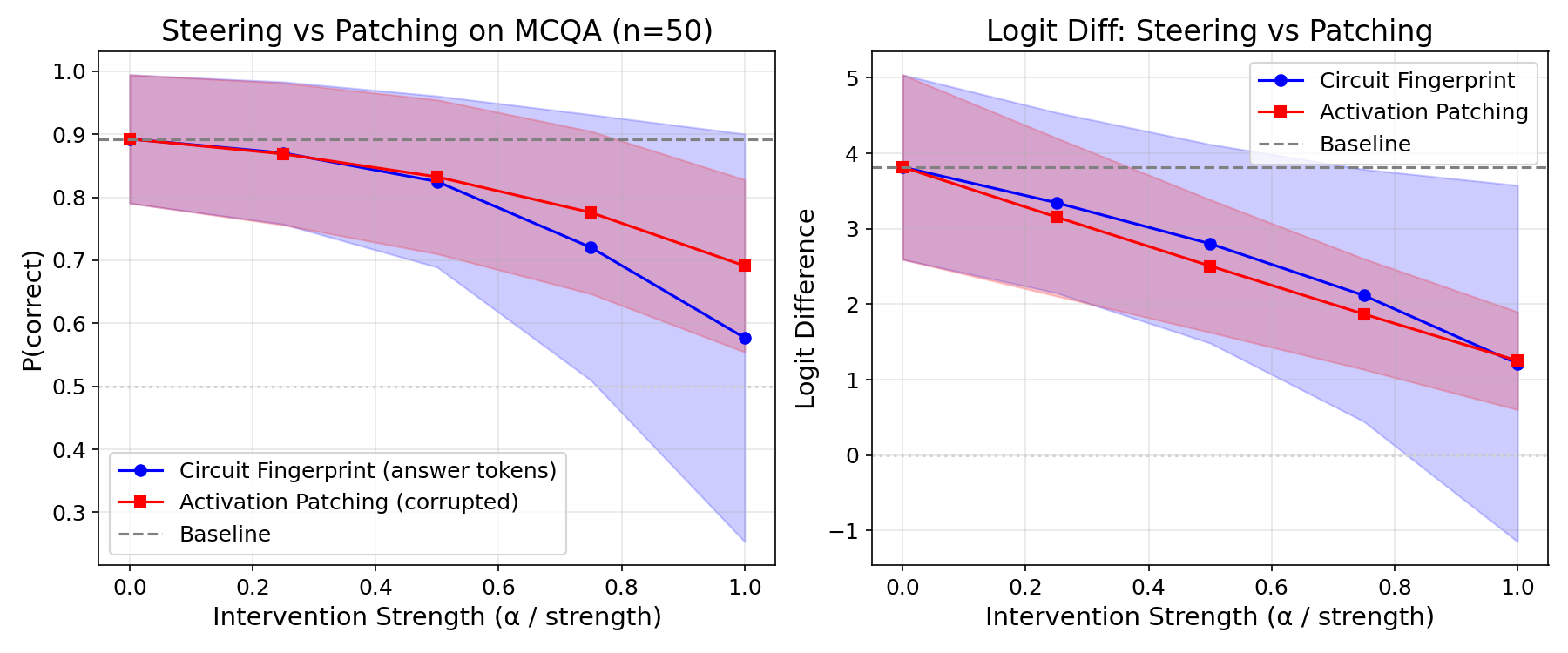}\\
    (b) Qwen2.5-0.5B
\end{minipage}

\vspace{0.5em}

\begin{minipage}{0.48\textwidth}
    \centering
    \includegraphics[width=\textwidth]{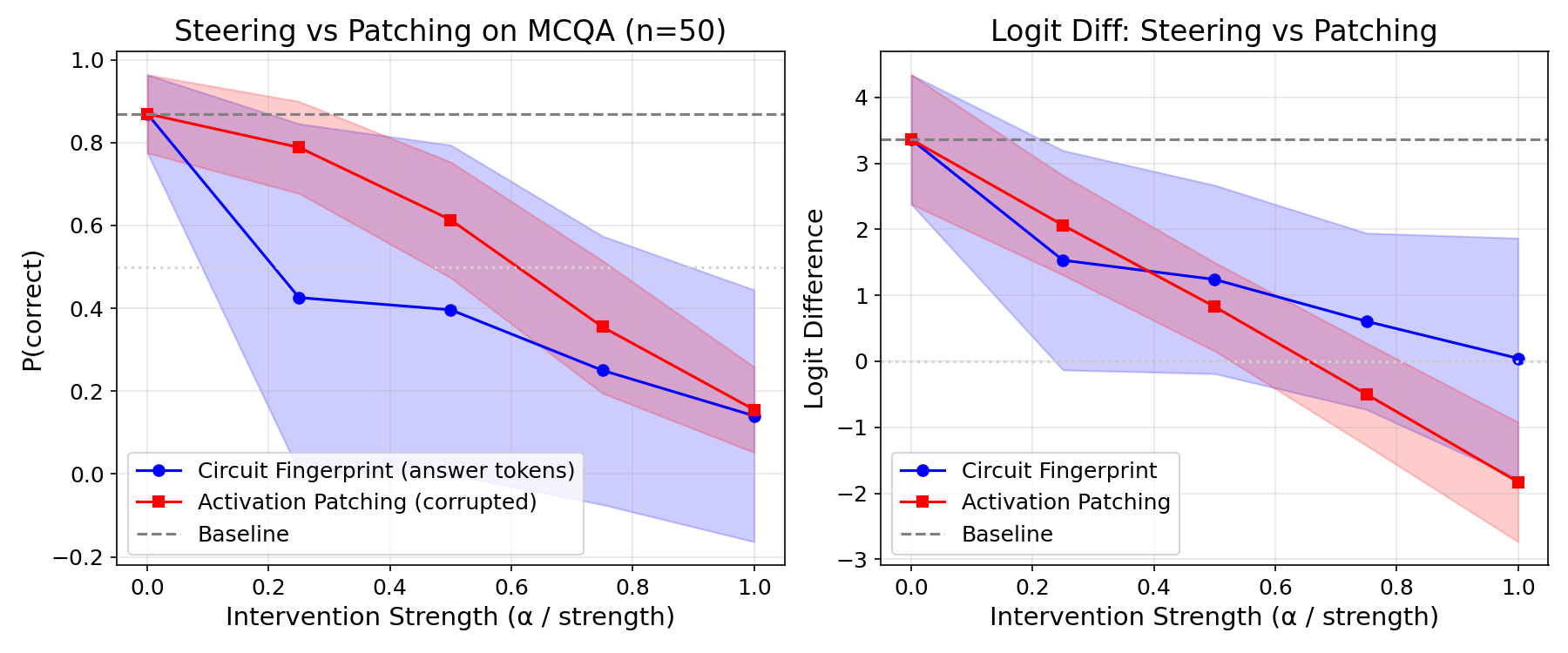}\\
    (c) Llama3.2-1B
\end{minipage}
\hfill
\begin{minipage}{0.48\textwidth}
    \centering
    \includegraphics[width=\textwidth]{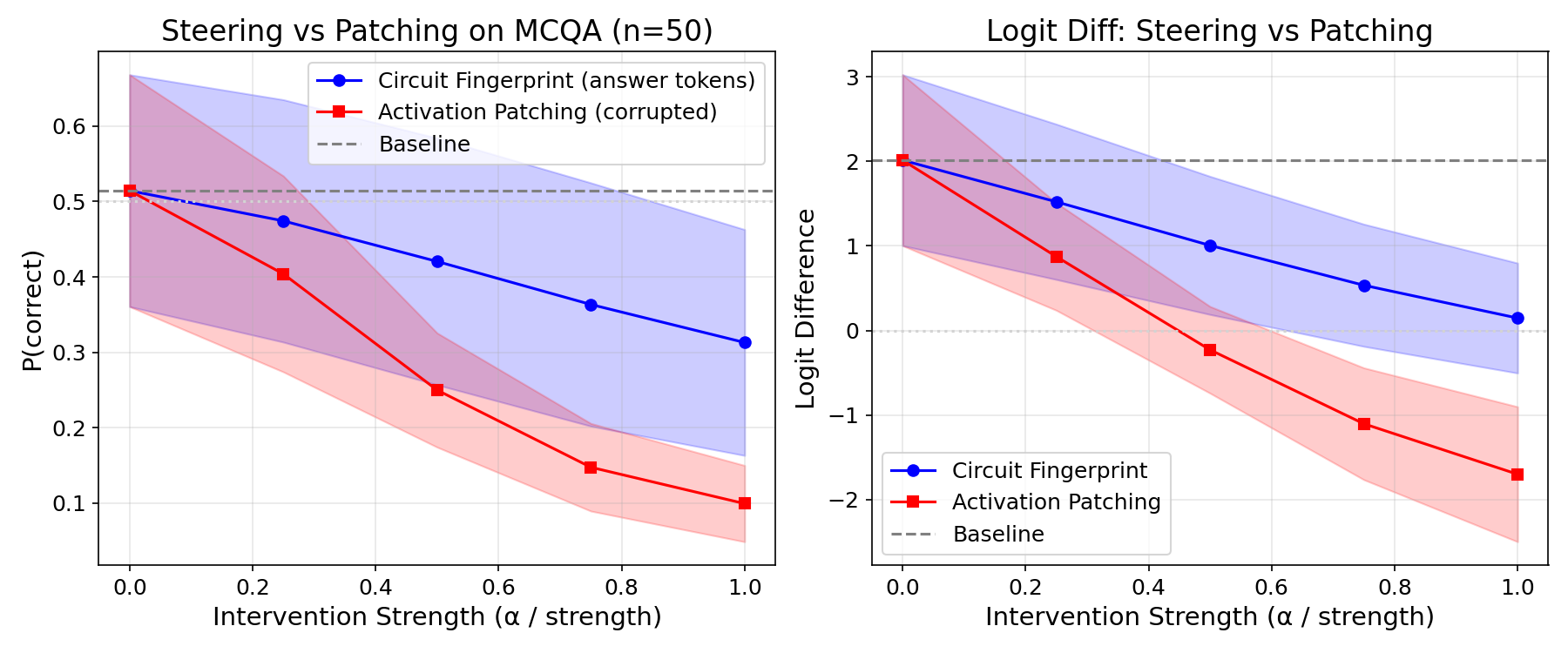}\\
    (d) OPT-1.3B
\end{minipage}
\caption{Activation Patching Comparisson for MCQA.}
\label{fig:all_models}
\end{figure}

\newpage
\section{Steering Class Vectors}
\label{app:steer_vec_res}

Table~\ref{tab:emotion_steering_detailed} presents per-emotion results. Geometric 
steering achieves higher classification accuracy on 3/5 emotions (joy, sadness, 
surprise), while consistently producing lower perplexity—indicating more fluent 
generations—without degrading factual accuracy (89.6\% vs 90.1\%).

\begin{table*}[h]
\centering
\caption{Per-Language Emotion Steering: Geometric Method vs Instruction Prompting}
\label{tab:emotion_steering_detailed}
\begin{tabular}{lccccc}
\toprule
\multirow{2}{*}{Emotion} & \multicolumn{2}{c}{Classification Acc.} & \multicolumn{2}{c}{Perplexity$\downarrow$} & Factual Acc. \\
\cmidrule(lr){2-3} \cmidrule(lr){4-5} \cmidrule(lr){6-6}
 & Baseline & Steered & Baseline & Steered & Steered \\
\midrule
Anger    & 55.1\% & 24.5\% & 10.50 & 11.12 & 96\% \\
Disgust  & 75.5\% & 67.3\% & 17.95 & 13.14 & 78\% \\
Joy      & 49.0\% & \textbf{87.8\%} & 15.76 & 12.93 & 100\% \\
Sadness  & 53.1\% & \textbf{89.8\%} & 18.09 & 13.95 & 81\% \\
Surprise & 32.7\% & \textbf{79.6\%} & 22.86 & 15.72 & 93\% \\
\midrule
Average  & 53.1\% & \textbf{69.8\%} & 17.03 & \textbf{13.37} & 90\% \\
\bottomrule
\end{tabular}
\end{table*}

\end{document}